\documentclass[manuscript, authorversion, table, review=false, timestamp=false, nonacm, screen]{acmart}
\usepackage{todonotes}

\usepackage{arydshln}
\usepackage{tablefootnote}
\usepackage[export]{adjustbox}
\usepackage{wrapfig}
\usepackage{graphicx}
\usepackage{algpseudocode} 
\usepackage[mathscr]{eucal}
\usepackage{amsmath}
\usepackage{mathtools}
\usepackage{multicol}
\usepackage{algorithm} 
\usepackage{bm}
\usepackage{hyperref}
\usepackage{booktabs}
\usepackage[applemac]{inputenc}
\usepackage{xparse}
\usepackage{xifthen}
\usepackage{mathtools}
\usepackage{multirow}
\usepackage{textcomp}
\usepackage{todonotes}
\usepackage{pifont}
\usepackage{soul}
\usepackage{color}
\usepackage{upgreek}

\usepackage{subfig}
\usepackage{graphicx}
\usepackage{footnote}
\makesavenoteenv{tabular}
\usepackage{lipsum}
\usepackage{xcolor}
\definecolor{anti-flashwhite}{rgb}{0.95, 0.95, 0.96}

\newcommand\blfootnote[1]{%
  \begingroup
  \renewcommand\thefootnote{}\footnote{#1}%
  \addtocounter{footnote}{-1}%
  \endgroup
}

\renewcommand{\textsl}{\textit}
\newcommand{\comm}[1]{}
\newcommand\numberthis{\addtocounter{equation}{1}\tag{\theequation}}
\NewDocumentCommand{\vect}{ O{} O{} m }{\mathbf{#3}\ifthenelse{\isempty{#1}}{}{^{(#1)}}\ifthenelse{\isempty{#2}}{}{_{#2}}}

\NewDocumentCommand{\mat}{ O{} O{} m }{\mathbf{#3}\ifthenelse{\isempty{#1}}{}{^{(#1)}}\ifthenelse{\isempty{#2}}{}{_{#2}}}

\NewDocumentCommand{\ten}{ O{} O{} m }{\pmb{\mathscr{#3}}\ifthenelse{\isempty{#1}}{}{^{(#1)}}\ifthenelse{\isempty{#2}}{}{_{#2}}}

\makeatletter
\newcommand{\thickhline}{%
    \noalign {\ifnum 0=`}\fi \hrule height 1pt
    \futurelet \reserved@a \@xhline
}

\DeclareMathOperator*{\argmin}{arg\,min}
\AtBeginDocument{%
  }

\setcopyright{none}
\copyrightyear{2022}
\acmYear{2022}
\acmDOI{none}

\acmConference[Pre-print. Under review.]{}{}{}
\acmPrice{15.00}
\acmISBN{978-1-4503-XXXX-X/18/06}

\begin{document}

\title[FedSPLIT: One-Shot Federated Recommendation System]{FedSPLIT: One-Shot Federated Recommendation System Based on Non-negative Joint Matrix Factorization and Knowledge Distillation}


\author{Maksim E. Eren}\blfootnote{This manuscript has been approved for unlimited release and has been assigned LA-UR-22-24122.}
\blfootnote{\textbf{Pre-print -- Under review.}}
\email{meren1@umbc.edu}
\affiliation{%
  \institution{University of Maryland, Baltimore County}
  \country{USA}
}

\author{Luke E. Richards}
\email{lurich1@umbc.edu}
\affiliation{%
  \institution{University of Maryland, Baltimore County}
  \country{USA}
}

\author{Manish Bhattarai}
\affiliation{%
  \institution{Theoretical Division, LANL}
  \country{USA}
}
\email{ceodspspectrum@lanl.gov}

\author{Roberto Yus}
\email{ryus@umbc.edu}
\affiliation{%
  \institution{University of Maryland, Baltimore County}
  \country{USA}
}

\author{Charles Nicholas}
\affiliation{%
  \institution{University of Maryland, Baltimore County}
  \country{USA}
}
\email{nicholas@umbc.edu}

\author{Boian S. Alexandrov}
\affiliation{%
  \institution{Theoretical Division, LANL}
  \country{USA}
}
\email{boian@lanl.gov}

\renewcommand{\shortauthors}{Eren, Luke, Bhattarai, Yus, Nicholas, and Alexandrov}

\begin{abstract}
  Non-negative matrix factorization (NMF) with missing-value completion is a well-known effective Collaborative Filtering (CF) method used to provide personalized user recommendations. However, traditional CF relies on the privacy-invasive collection of users' explicit and implicit feedback to build a central recommender model. One-shot federated learning has recently emerged as a method to mitigate the privacy problem while addressing the traditional communication bottleneck of federated learning. In this paper, we present the first unsupervised one-shot federated CF implementation, named FedSPLIT, based on NMF joint factorization. In our solution, the clients first apply local CF in-parallel to build distinct client-specific recommenders. Then, the privacy-preserving local item patterns and biases from each client are shared with the processor to perform joint factorization in order to extract the global item patterns. Extracted patterns are then aggregated to each client to build the local models via knowledge distillation. In our experiments, we demonstrate the feasibility of our approach with standard recommendation datasets. FedSPLIT can obtain similar results than the state of the art (and even outperform it in certain situations) with a substantial decrease in the number of communications.

\end{abstract}

\ccsdesc[500]{Security and privacy~Privacy protections}
\ccsdesc[500]{Computing methodologies~Dimensionality reduction and manifold learning}
\ccsdesc[500]{Computing methodologies~Non-negative matrix factorization}

\keywords{privacy, non-negative matrix factorization, one-shot, federated learning, recommendation system}

\maketitle

\section{Introduction}

Machine learning (ML) has grown in popularity in the past decade as methods matured with the growing availability of computation and large-scale data. One of the domains in which ML has been successfully used is that of recommender systems. ML-powered recommender systems have become an integral part of our lives in a broad range of applications, such as selecting movies, books, music, and other merchandise in e-commerce. While users enjoy enhanced and personalized user experience with recommender systems, companies improve their sales and customer loyalty \cite{lu2012recommender, chen2004impact}. Despite the benefits, this success has resulted in important ethical challenges~\cite{milano2020recommender}. Among those, the evergrowing collection of individuals' data to provide personalized and sophisticated recommendations has resulted in challenges to our privacy.

Traditional recommender ML methods rely on privacy-invasive collection of user data to train central models at the processors (servers, or data processing organization), resulting in loss of control and ownership of the private user data. In today's world, users' expectations concerning privacy are changing with the growing awareness of privacy, and new more stringent data privacy regulations are being introduced globally. For instance, the European General Data Protection Regulation (GDPR) has implications on how recommendation systems collect and handle user data~\cite{krebs2019tell}. This movement drives the need for future recommendation systems to take the preservation of privacy into account to comply with the regulations and help in maintaining consumer trust and loyalty. Recommendation systems based on federated learning are an attractive solution to minimize the private data that is transmitted to the processors~\cite{yang2020federated}. While federated learning alone is also vulnerable to potential privacy leakage~\cite{liu2022threats}, it contributes to minimizing the amount of individuals' data that flows to processors by performing part of the training process at the client. However, one of the core challenges of this approach is the expensive communication cost between the clients and the server~\cite{yang2020federated}.

The communication challenge of federated learning has an impact on federated recommender systems that might jeopardize their wide adoption. Several federated recommenders based on Collaborative Filtering (CF) implementations have been previously proposed \cite{ammad2019federated, flanagan2020federated}. Such systems, which identify and filter potential future user interests by collaboratively learning from the past preferences of many users \cite{goldberg1992using, resnick1997recommender}, are based on iterative optimization techniques requiring multiple communications between the server and its clients at each iteration.
Federated methods that require multi-round client participation suffer communication bottlenecks due to potential limited data plans of users, potentially slow and unreliable network connections, and cost of cryptographic protocols \cite{li2020federated, Guha2019OneShotFL}. Also, the higher the number of communications between clients and server, the higher the risk for potential attacks to intercept model updates that could lead back to individuals' data. Thus, communication efficiency in federated learning has been an active field of study \cite{li2020federated, mcmahan2016federated, konevcny2015federated, konevcny2016federated, smith2017federated, wang2019matcha, luping2019cmfl}. In particular, different approaches have been proposed to improve the efficiency of communications in federated CF \cite{chen2018federated, jalalirad2019simple, singhal2021federated, lin2020fedrec}. While these approaches lower the communication complexity, they still expect multi-round client participation. This presents challenges in the real world as clients may leave the training process at any moment \cite{Li2021PracticalOF} or have different number of data instances \cite{mcmahan2017communication}. In response, an emerging field of \textit{one-shot federated learning} addresses the communication problem by performing \textit{only a single round of communication} (or a fixed number of few communications) between each client and the processor per training session, as opposed to per iteration \cite{Guha2019OneShotFL, Zhou2020DistilledOF, Li2021PracticalOF, yurochkin2019bayesian, kasturi2020fusion, shin2020xor, guha2018model}. While the proposed one-shot federated methods significantly reduce the cost of communication, work thus far only focused on supervised and semi-supervised classification problems such as language and vision tasks.

In this paper, we introduce what is, to the best of our knowledge, the first implementation of one-shot federated learning for CF and recommendations. Our approach, named FedSPLIT, is an unsupervised communication-efficient (for the number of communications) federated method for providing privacy to groups of users or set of organizations (clients). In our one-shot federated setup, we achieve a \textit{single pair of communication}, after the initial small-sized communication to calculate the global mean, between the server and its clients for the entire training process by first modeling local client-specific CF in parallel. Specifically, our local CF is based on Non-negative Matrix Factorization (NMF) from users' explicit feedback data \cite{su2009survey, koren2009matrix, Hug2020}. The extracted latent factors for items from each client along with the item biases are then shared with the processor to extract the global item patterns via joint-factorization. Global item patterns and biases are then aggregated to each client to perform knowledge distillation to improve the recommendation capability of the distinct local models. Since the shared data between the clients and server is with respect to the items, the added benefit of FedSPLIT is that the user based information is abstracted, where the processor does not know which user is part of which group (client) or how many users are in the group. During our experiments, we showcase the feasibility of our approach on the MovieLens 100K and MovileLens 1M datasets \cite{harper2015movielens} and show that FedSPLIT achieves similar root-mean-square error (RMSE) with a single communication as compared to the state-of-the-art federated CF with multi-round communication. We also perform an empirical privacy audit of our method by analyzing how much outside knowledge an adversary would need to reconstruct the dataset of a group or organization. In summary, our main contributions include:
\begin{itemize}
    \item Introducing the first implementation of a one-shot federated collaborative filtering method for performing recommendations from user explicit feedback.
    \item Developing a one-shot federated learning method where user information is abstracted.
    \item Demonstrating that NMF with joint factorization and knowledge distillation improves the recommendation capability of groups or organizations.
    \item Showcasing that our approach achieves similar RMSE with a single communication as compared to the state-of-the-art federated CF.
    \item Presenting an empirical privacy evaluation of our method to analyze potential reconstruction attacks via inference.
\end{itemize}

The rest of the paper is organized as follows. We give a summary of relevant prior work in Section \ref{sec:relevant_work}. We describe NMF, NMF with missing-value completion for CF, and FedSPLIT in sections \ref{sec:nmf}, \ref{sec:masked_nmf}, and \ref{sec:split}, respectively. We also provide an empirical summary of privacy by design considerations and potential limitations of our approach in Section \ref{sec:privacy_by_design}. Before we showcase our experiment results, the dataset and the experimental setup are described in Section \ref{sec:dataset_and_setup}. We then demonstrate that federated learning improve the predictive performance of local CF in Section \ref{sec:performance_analysis}, before comparing our results to the baseline federated CF methods in Section \ref{sec:baseline_comparisions}. The privacy audit is given in Section \ref{sec:privacy_audit}. Before concluding, we list some potential areas of future work in Section \ref{sec:future_work}.

\section{Related Work}
\label{sec:relevant_work}

In this section, we provide a brief overview of relevant work on federated learning, and specifically the prior work that looked at federated learning in the context of federated CF and the optimization of communication.

\subsection{Federated Learning}

Federated learning \cite{mcmahan2017communication} formalized the concept of learning local models and then aggregating them to create a centralized global model in the pursuit of sharing knowledge derived from data. Since then there has been a growing level of research in furthering this method \cite{li2020federated}. Works since have examined more efficient communication with reducing the number of communications or the size of the communicated data \cite{mcmahan2016federated, konevcny2015federated, smith2017federated, wang2019matcha, luping2019cmfl, konevcny2016federated}, ensuring privacy of participants \cite{ying2020shared, basu2020privacy, 10.1145/3460231.3474262}, and extending to recommendation systems \cite{tan2020federated, zhang2021vertical, 10.1145/3460231.3474262, ammad2019federated, chen2018federated, jalalirad2019simple, flanagan2020federated, singhal2021federated}. A major caveat of these methods, especially those within the recommendation system space, is the need for consistent and frequent communication of gradients and model updates. Communication optimization work in federated learning \cite{khan2021payload, mcmahan2016federated, konevcny2015federated, smith2017federated, wang2019matcha, luping2019cmfl, konevcny2016federated} has gained much more attention as these systems were implemented in real world scenarios and constraints. Our work is most similar to the concept of one-shot federated learning \cite{Guha2019OneShotFL} where the goal is to limit communication to one iteration to aggregate knowledge across models. A handful of approaches have examined methods to aggregate model knowledge through distillation \cite{Zhou2020DistilledOF, Li2021PracticalOF}, data distribution modeling for synthetic data generation \cite{Zhou2020DistilledOF, kasturi2020fusion}, model selection for aggregation \cite{guha2018model}, probabilistic aggregation \cite{yurochkin2019bayesian} and addressing privacy-preserving aggregation for non-IID data \cite{shin2020xor}. Methods have been evaluated in supervised and semi-supervised image and text classification domain, and to our knowledge our work introduces the first one-shot federated learning method for unsupervised recommendation. 

Work in federated recommendation systems, specifically matrix factorization \cite{ying2020shared,YANG2021106946,yang2021practical} attempts to solve a problem that is similar to our work within collaborative filtering, with certain users being present in distributed matrices and items being shared. 
However, they focus on a global federated learning by sharing gradients with a secret sharing mechanism with the central server. Our work is in a similar vein as \cite{alsulaimawi2020non} in which NMF is used as a local privacy-filter within a federated learning paradigm. Their approach involves learning a local latent representation that allows for a utility classification while protecting sensitive attributes from being reconstructed. Within our approach, we also attempt to learn a latent representation for the utility of collaborative filtering rather than classification. 

\subsection{Private Machine Learning and Collaborative Filtering}

Private ML attempts to create models that do not leak information on the original data that the model was trained on \cite{gong2020survey}. This is an ever prevalent issue within federated learning as there are typically frequent network communications between devices and server which leaves user data vulnerable \cite{gao2020privacy}. Algorithms attempt to mitigate membership inference attacks \cite{shokri2017membership}, being able to determine if a data point was used in training, and inversion attacks \cite{fredrikson2015model}, and being able to reproduce the dataset the model was trained on. Various methods have been introduced to accomplish these tasks such as homomorphic encrypted models \cite{sun2018private} and through differential privacy algorithms \cite{abadi2016deep}. Complete homomorphic encryption in matrix factorization has been explored \cite{kim2016efficient} outside of the federated learning environment and within it for sending data to the central server \cite{chai2020secure}. Differential privacy has been explored to provide protection in the federated collaborative filtering context \cite{10.1145/3460231.3474262} and outside of it with count sketches as an approximation \cite{balu2016differentially}. Methods have proposed not retaining data sent from users and only retaining online-updated item matrices to preserve privacy \cite{vallet2014matrix}. Our work shares a similar goal of not retaining data, but does not require individuals to share data points and does not store latent matrices that are updated in an online fashion. 

Within our work we examine the privacy of our newly introduced method. The majority of methods within this space use gradient information sent to the central server on the weights of the model. However prior work \cite{zhu2019deep} has shown a method to recover private learning data from these exchanged gradients. This attack is done by utilizing the local version of the model and using the shared weights to optimize random inputs to the original data to produce the same gradients. In response, works have examined how to integrate homomorphic encryption for data sharing \cite{zhang2021vertical, chai2020secure} and model gradients \cite{basu2020privacy}. Prior work in federated collaborative filtering similarly evaluated the privacy of federated matrix factorization with gradient sharing \cite{gao2020privacy}. In a similar vein as prior works, they were able to reconstruct the original data from unprotected gradients. Within our work we forgo sharing gradients entirely and evaluate the privacy risk of sharing latent components. As well, we explore the dynamics of sharing data within a group setting for recommendation \cite{o2001polylens} and when this grouping of user data introduces potential group data reconstruction attacks.

\section{Method}
\label{sec:method}

The core ideology in our method is that we want to jointly learn the patterns of data that is \textit{split} into separate matrices, hence the name Fed\textit{SPLIT}. On the client-side, FedSplit uses a Collaborative NMF for missing-value completion, an effective CF method for performing recommendations from users' explicit feedback data \cite{su2009survey, koren2009matrix}, while the server-side utilizes standard NMF for joint factorization. Before introducing FedSplit and privacy and practical design considerations, we begin this section with a summary of the standard NMF and the Collaborative NMF.

\subsection{Non-negative Matrix Factorization (NMF)}
\label{sec:nmf}

NMF is an unsupervised learning method based on low-rank approximation that performs dimensionality reduction. NMF approximates a given observation matrix $\mat{X}\in \mathbb{R}_{+}^{n \times m}$ with non-negative entries, as a product of two non-negative matrices, i.e., $\mat{X} \approx \mat{W}\mat{H}$, where $\mat{W} \in \mathbb{R}_{+}^{n\times k}$, and $\mat{H} \in \mathbb{R}_{+}^{k \times m}$, and usually $k\ll m, n$. Here, $n$ is the number of samples, $m$ is the number of features, and the small dimension $k$ is the low-rank of the approximation. We perform this factorization via a non-convex minimization with non-negativity constraint, utilizing the multiplicative updates algorithm \cite{lee1999learning}, and  Frobenius norm as distance metric, with an objective function:
\begin{align*}
\label{eqn:nmf_min}
    \underset{\mat{W} \in \mathbb{R}_{+}^{n\times k}, \, \mat{H} \in \mathbb{R}_{+}^{k \times m}}{\operatorname{min}} \, ||\mat{X} - \mat{W}\mat{H}||_{F}^{2} \numberthis 
\end{align*}
which allows NMF to be treated as a Gaussian mixture model \cite{fevotte2009nonnegative}. In Equation \ref{eqn:nmf_min}, the factors $\mat{W}$ and $\mat{H}$ are the solution of the optimization problem, estimated via alternative updates, with update rules in Equations \ref{eqn:NMF_updates_W} and \ref{eqn:NMF_updates_H} respectively, and the performance of the minimization is evaluated by the relative reconstruction error in Equation \ref{eqn:rel_error}:

\begin{multicols}{3}
\noindent
\begin{align}
\label{eqn:NMF_updates_W}
\mat{W} =\mat{W}\frac{\mat{X}\mat{H}^T}{\mat{W} \mat{H} \mat{H}^T}
\end{align}
\begin{align}
\label{eqn:NMF_updates_H}
\mat{H} =\mat{H}\frac{\mat{W}^T\mat{X}}{\mat{W}^T \mat{W} \mat{H}}
\end{align}
\begin{align}
\label{eqn:rel_error}
\text{Relative Error} =\frac{||\mat{X}-\mat{W}\mat{H}||_F^2}{||\mat{X}||_F^2}
\end{align}
\end{multicols}

\subsection{Collaborative Non-negative Matrix Factorization (CNMF)}
\label{sec:masked_nmf}

While NMF, as summarised in Section \ref{sec:nmf}, is an effective method for extracting the meaningful latent features, the updates are done with respect to Equation \ref{eqn:nmf_min} such that the distance between $\mat{X}$ and approximation $\mat{W}\mat{H}$ is minimized. This distance minimization includes the zero entries in $\mat{X}$ such that the approximation $\mat{W}\mat{H}$ is pushed towards zero for those entries. However, in recommendation systems, the user only interacts with a small fraction of the items, which leads to a highly sparse matrix $\mat{X}$. Therefore, the standard NMF minimization does not work for CF because the zeros, or the missing-values, in $\mat{X}$ are the future recommendations we want to perform. To this end, our optimization needs to be done only with respect to the non-zero entries in $\mat{X}$ instead. Estimating the missing data value in this way, i.e. users' future recommendations, is called matrix completion problem \cite{xu2012alternating}. For this, we use the modified version of the Collaborative NMF algorithm presented in the \textit{Surprise} package \cite{Hug2020}, and call it CNMF for simplicity. We modify Surprise with non-negative projections to ensure the non-negativity in the latent factor matrices, and adopt its bias terms to the federated learning scheme (Section \ref{sec:split}).

Consider $\mat{X}\in \mathbb{R}_{+}^{n \times m}$ to be a matrix of movie ratings from $n$ users for $m$ movies where each user $i$ rated a subset of $m$ items, such that a non-zero entry $r^{i,j}=\mat{X}_{i,j}$ is the rating from user $i$ for movie $j$. The user profile matrix is given by $\mat{W}\in \mathbb{R}_{+}^{n \times k}$ and the movie profile matrix is given by $\mat{H}\in \mathbb{R}_{+}^{k \times m}$. The future rating of user $i$ for movie $j$ is predicted with:
\begin{equation}
\label{eqn:predict_NMF}
\hat{\mat{X}}_{i,j} = \mat{W}_{i,:}\mat{H}_{:,j}
\end{equation}

For a more compelling matrix approximation in recommendation systems, we also consider the bias terms $b_W \in \mathbb{R}^{n}$ and $b_H\in \mathbb{R}^{m}$ to remove the bias given by users or bias for an item. Here $b_H$ describes how well an item is rated compared to the average across all the items before accounting for the interaction between user and item. Similarly, $b_W$ describes the user's tendency to provide better/worse ratings compared to average. In addition, we consider the group bias. If we let $\text{\textit{nnz}}(\mat{X})$ represent a vector of
non-zero entries (ratings) in $\mat{X}$, then group bias is calculated as follows:
\begin{equation}
\label{eqn:mean_bias}
\mu = \dfrac{1}{|\text{\textit{nnz}}(\mat{X})|} \cdot \sum^{|\text{\textit{nnz}}(\mat{X})|}_{i} \text{\textit{nnz}}(\mat{X})_i
\end{equation}

\noindent such that $\mu$ is simply the average of the ratings. Then, Equation \ref{eqn:predict_NMF} can be reformulated for predicting the rating from user $i$ for movie $j$ as shown in the Equation \ref{eqn:predict_MNMF}, and the performance can be measured with RMSE (Equation \ref{eqn:RMSE}):

\begin{multicols}{2}
\noindent
\begin{align}
\label{eqn:predict_MNMF}
\hat{\mat{X}}_{i,j} = \mat{W}_{i,:}\mat{H}_{:,j} + b_{W_i} + b_{H_j} + \mu
\end{align}
\begin{align}
\label{eqn:RMSE}
\text{RMSE} = \frac{||\mat{X}-\hat{\mat{X}}||_F^2}{\sqrt{mn}} 
\end{align}
\end{multicols}

Now, to estimate the factors $\mat{W}$, $\mat{H}$ and bias components $b_H$ and $b_W$, we rewrite the objective function (Equation \ref{eqn:nmf_min}):

\begin{align*}
\label{eqn:optimize}
   \argmin_{\mat{W} \in \mathbb{R}_{+}^{n\times k}, \, \mat{H} \in \mathbb{R}_{+}^{k \times m}, \, b_W, \, b_H} ||\mat{\mat{X}}_{i,j}-(\mat{W}_{i,:}\mat{H}_{:,j}+b_{W_i}+b_{H_j}+\mu)||^2_F + \alpha||\mat{W}||^2_F + \beta ||\mat{H}||^2_F+ \gamma||b_W||^2_F + \delta ||b_H||^2_F \numberthis
\end{align*}

\noindent where $i$ and $j$ correspond only to the non-zero coordinates (ratings) such that $\mat{X}_{i,j} > 0$ for each $i$ and $j$, and $\alpha$, $\beta$, $\gamma$ and $\delta$ are the regularization parameters for $\mat{W}, \mat{H}, b_W$ and $b_H$, respectively. The above regularization is based on Tikhonov regularization \cite{golub1999tikhonov} for regularizing the ill posed problems to obtain higher prediction accuracy. We solve Equation~\ref{eqn:optimize} with gradient descent updates to estimate  biases $b_W$ and $b_H$ with learning rates $\eta_W$ and $\eta_H$, respectively (Equations~\ref{eqn:bias_W_updates} and~\ref{eqn:bias_H_updates}) . We also estimate the factors $\mat{W}$ and $\mat{H}$ with closed form update expressions (Equations~\ref{eqn:masked_W_updates} and~\ref{eqn:masked_H_updates}). The updates for these parameters are performed alternately until convergence. The update steps are:

\begin{multicols}{2}
\noindent
\begin{align}
\label{eqn:bias_W_updates}
b_W= b_W+\eta_W \sum_{j=1}^m(\mat{err}_{:,j}-\gamma*b_W)
\end{align}
\begin{align}
\label{eqn:bias_H_updates}
b_H= b_H+\eta_H \sum_{i=1}^n(\mat{err}_{i,:}-\delta*b_H) 
\end{align}
\end{multicols}

\begin{multicols}{2}
\noindent
\begin{align}
\label{eqn:masked_W_updates}
\mat{W} =\mat{W}\frac{\mat{X}\mat{H}^T}{\hat{\mat{X}}\mat{H}^T+\alpha \mat{W}} \numberthis
\end{align}
\begin{align}
\label{eqn:masked_H_updates}
\mat{H} =\mat{H}\frac{\mat{W}^T\mat{X}}{\mat{W}^T\hat{\mat{X}}+\beta \mat{H}}
\end{align}
\end{multicols}

\noindent  where $\hat{\mat{X}}_{i,j}$ is the estimation and calculated via Equation \ref{eqn:predict_MNMF}, and $\mat{err}_{i,j}=\mat{X}_{i,j}-\hat{\mat{X}}_{i,j}$.

\subsection{FedSPLIT: One-shot Federated Collaborative Filtering}
\label{sec:split}

\begin{figure}[htb]
  \includegraphics[width=\columnwidth]{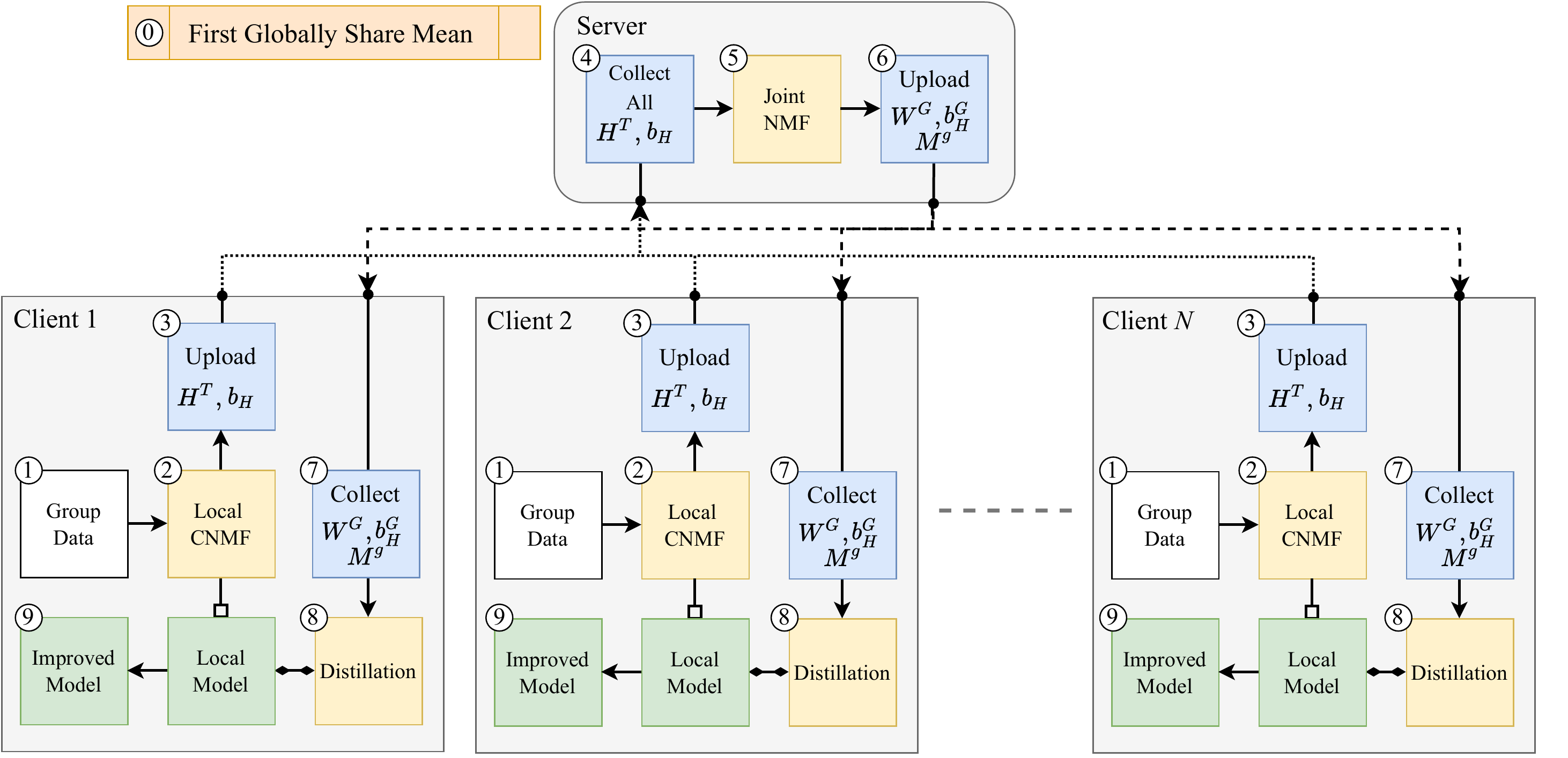}
  \caption{Summary of the FedSPLIT methodology. Points 0, 1, 2, and 3 are described in Section \ref{sec:step_1}. Points 4, 5, and 6 are described in Section \ref{sec:step_2}. Points 7 and 8 are described in Section \ref{sec:step_3}. Finally, point 9 is described in Section \ref{sec:step_4}. Superscript $G$ is short for $Global$.\label{fig:fedsplit}}
\end{figure}

\begin{algorithm}[htb]
\setstretch{1.10}
    \caption{FedSPLIT(Clients, Processor) \label{algorithm:FedSPLIT_alg}} 
    \begin{algorithmic}[1]
    
    \State $\mu^{Global}$ = Processor.mean \Comment{\textbf{Communication}, Global mean}
    
    \For {client in Clients} \Comment{In parallel for $N$ Clients}
        \State client.mean = $\mu^{Global}$ \Comment{Save global mean rating}
        \State client.$\mat{W}$, client.$\mat{H}$, client.$b_W$, client.$b_H$ = CNMF(client.$\mat{X}$, client.mean) \Comment{CNMF to extract local patterns}
    
    \EndFor
    
    \State Processor $\leftarrow$ \textbf{Download} $(\mat{H})^T$ and $b_H$ from each $N$ client \Comment{\textbf{Communication}, where superscript $T$ is the transpose}
    \State Processor.$b_{H}^{Global}$ = $\dfrac{1}{N}\sum^{N}_{g}b_{H}^{g}$ \Comment{Global item biases, where $b_{H}^{g}$ is from the $g$th client, and $0 \leq g \leq N$}
    \State Processor.$\mat{X}$ = $[(\mat{H}^{1})^T, (\mat{H}^{2})^T, \cdots, (\mat{H}^{N})^T]$ \Comment{Where $(\mat{H}^{g})^T$ is from $g$th client}
    \State Processor.$\mat{W}^{Global}$, Processor.$[\mat{M}^{1}, \mat{M}^{2}, \cdots, \mat{M}^{N}]$ = NMF(Processor.$\mat{X}$) \Comment{Standard NMF to extract global patterns}
    \State \textbf{Upload} $\mat{W}^{Global}$, $\mat{M}^{g}$, and $b_{W}^{Global}$ $\rightarrow$ Clients \Comment{\textbf{Communication}}
    
    \For {client in Clients} \Comment{In parallel for $N$ Clients}
        \State client.$\mat{W}^{*}$ =   client.$\mat{W}$ $\circ$ (client.$\mat{M}^{g})^T$ \Comment{Knowledge distillation}
    \EndFor 
    \end{algorithmic} 
\end{algorithm}

Now that we have summarized NMF and Collaborative NMF, we introduce our one-shot federated recommendation system named FedSPLIT in this section, which is also summarized in Figure \ref{fig:fedsplit} and Algorithm \ref{algorithm:FedSPLIT_alg}. We describe our method in four steps.

\subsubsection{Step 1 - Clients (Local CF via Collaborative NMF)}
\label{sec:step_1}

Let us consider a set of $N$ clients each with their local data $\mat{X}^{g} \in \mathbb{R}_{+}^{n^g \times m}$, where $g$ is in range $0 \leq g \leq N$, and non-zeros in $\mat{X}^{g}$ are the ratings for $m$ items (e.g., movies) by $n^g$ users belonging to the group $g$. The first step in FedSPLIT involves a small-sized pair of communications between each client and the server, where each client sends the group bias $\mu^{g}$ (Equation \ref{eqn:mean_bias}) to the server and receives the global bias $\mu^{Global}$ (Equation~\ref{eqn:globalbias}, Line 1 in Algorithm \ref{algorithm:FedSPLIT_alg}) in order to put global bias into consideration of local CNMF:
\begin{align*}
\label{eqn:globalbias}
    \mu^{Global} = \dfrac{\mu^{1} + \mu^{2} + \cdots + \mu^{g} + \cdots + \mu^{N}}{N} \numberthis
\end{align*}

Next, using $\mu^{Global}$ and $\mat{X}^{g}$, each of the $N$ clients trains local CF models in parallel using the Collaborative NMF approach as described in Section \ref{sec:masked_nmf} to obtain local patterns $\mat{W}^{g} \in \mathbb{R}_{+}^{n^g \times k^{g}}$ and $\mat{H}^{g} \in \mathbb{R}_{+}^{k^{g} \times m}$, and biases $b_{W}^{g}$ and $b_{H}^{g}$ (Lines 2 through 5 in Algorithm \ref{algorithm:FedSPLIT_alg}). At this point each client has a local CF model. However, we would like to utilize federated learning to improve the performance of these local models. Therefore, we proceed with our method where each client next uploads their $(\mat{H}^{g})^T$ and $b_{H}^{g}$ to the server (the superscript $T$ represents the matrix transpose).

\subsubsection{Step 2 - Server (Joint Factorization via NMF)}
\label{sec:step_2}

Once the processor receives the local item factors and the item biases from each client, it first forms the global $\mat{X} \in \mathbb{R}_{+}^{m \times (k^1 + k^2 + \cdots + k^{g} + \cdots + k^N)}$ ( Equation~\ref{eqn:form_X}) and calculates the global item bias (Equation~\ref{eqn:form_bias}) (Line 7 and 8 in Algorithm \ref{algorithm:FedSPLIT_alg}):

\begin{multicols}{2}
\noindent
\begin{align}
\label{eqn:form_X}
\mat{X} = [(\mat{H}^{1})^T, (\mat{H}^{2})^T, \cdots, (\mat{H}^{g})^T, \cdots, (\mat{H}^{N})^T]
\end{align}
\begin{align}
\label{eqn:form_bias}
b_{H}^{Global} = \dfrac{b_{H}^{1} + b_{H}^{2} + \cdots + b_{H}^{g} + \cdots + b_{H}^{N}}{N}
\end{align}
\end{multicols}

Because $\mat{X}$ is the concatenation of each $(\mat{H}^{g})^T$ from the clients, we can perform joint factorization to identify the global item patterns. However, since the joint factor matrix $\mat{X}$ is dense, we can no longer perform updates based on matrix completion as described in Section \ref{sec:masked_nmf}. Instead, we apply standard NMF as described in Section \ref{sec:nmf} to obtain the factor matrices $\mat{W}^{Global} \in \mathbb{R}_{+}^{m \times K}$ and $\mat{H}^{Global} \in \mathbb{R}_{+}^{K \times (k^1 + k^2 + \cdots + k^{g} + \cdots + k^N)}$ (Line 9 in Algorithm \ref{algorithm:FedSPLIT_alg}). Here the objective is to estimate common item patterns across the $N$ groups. We let $\mat{M}^g$ represent a slice from $\mat{H}^{Global}$ belonging to group $g$:
\begin{align*}
\label{eqn:m_g}
    \mat{M}^g = \mat{H}^{Global}_{:, k^1 + k^2 + \cdots + k^{g-1}: k^{g} + k^{g+1} \cdots + k^{N}} \numberthis
\end{align*}

\begin{figure}[htb]
  \includegraphics[scale=0.6]{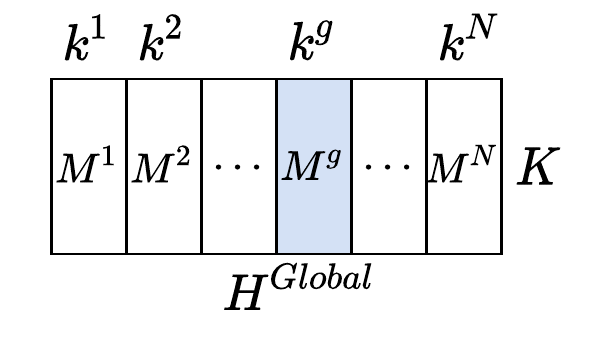}
  \caption{Demonstration of how each $\mat{M}^{g}$ is obtained from $\mat{H}^{Global}$ (Equation \ref{eqn:m_g}).}
  \label{fig:Mg}
\end{figure}

We also demonstrate how each $\mat{M}^{g}$ is obtained in Figure \ref{fig:Mg}. Note that, since the server does not have access to clients' latent feature matrix for user patterns $\mat{W}^g$, and the user biases $b^{g}_{W}$, the server cannot directly re-construct the private data $\mat{X}^{g}$ of group $g$. Hence, providing the server with only $(\mat{H}^{g})^T$, $b_{H}^{g}$, and $\mu^{g}$ allows minimizing the shared data and providing a level of privacy to the groups. Finally, the server broadcasts $\mat{M}^g$ to each client $g$ along with $\mat{W}^{Global}$ and $b_{H}^{Global}$.

\subsubsection{Step 3 - Clients (Knowledge Distillation)}
\label{sec:step_3}

Once the client $g$ receives the global patterns $\mat{M}^g$ and $\mat{W}^{Global}$, and the global item bias $b_{H}^{Global}$, the next step is to perform knowledge distillation to transfer the global patterns to the local user factor matrix $\mat{W}^{g}$. The goal of this step is to improve the recommendation capability of the local CF model. This is simply done with a single step using $\mat{M}^g$ as follows (Line 11 through 13 in Algorithm \ref{algorithm:FedSPLIT_alg}):
\begin{align*}
\label{eqn:distilation}
    \mat{W}^{*g} = \mat{W}^{g} (\mat{M}^{g})^T \numberthis
\end{align*}

\subsubsection{Step 4 - Clients (Local Recommendations)}
\label{sec:step_4}

At this point each of the $N$ clients has an improved client-specific CF model. Using these local distinct models each client can perform rating estimation for recommendations. For example, a rating of movie $j$ for user $i$ in group/client $g$ is now estimated as follows:

\begin{align*}
\label{eqn:predict_fed_split}
\hat{\mat{X}}^{g}_{i,j} = \mat{W}^{* g}_{i,:}(\mat{W}^{Global})^{T}_{:,j} + b_{W_i}^{g} + b_{H_j}^{Global} + \mu^{Global} \numberthis
\end{align*}

\subsubsection{Communication Summary}
FedSPLIT first performed a setup communication of sharing the average ratings $\mu^{g}$ to calculate $\mu^{Global}$, which is a small-sized communication (Section \ref{sec:step_1}). After the initial setup, there was a single round of communication between the clients and the server (Section \ref{sec:step_2}). Therefore, our approach does not do numerous communications like the iterative federated learning methods for recommendation system \cite{Singhal2021FederatedRP, DU2021107700, zhang2021vertical, YANG2021106946, chai2020secure, wu2021fedgnn, yang2021practical, lin2020fedrec, ying2020shared, 10.1145/3460231.3474262, khan2021payload, ammad2019federated, chen2018federated, jalalirad2019simple, flanagan2020federated, singhal2021federated}. Instead, FedSPLIT performs one-shot communication after the initial small-sized setup.

\subsection{Privacy by Design and Practical Implementation Considerations}
\label{sec:privacy_by_design}

The goal of FedSPLIT is to increase the feasibility of federated learning for recommendation by reducing the number of rounds of communication between the clients and the server. Federated learning helps in mitigating privacy challenges of ML-based recommender systems by minimizing the data shared with the processor. In this section, we take a closer look at how FedSPLIT contributes to preserving individuals' privacy and discuss potential limitations of our approach.

\subsubsection{Control and flow of data:}
\label{subsec:data_flow_control}
In FedSPLIT, users have local control of their data $\mat{X}^{g}$ within a group, which extends to the context of collaborating organizations where each organization has local control of their data \cite{o2001polylens, baltrunas2010group, ghosh2020efficient}. In addition to their local data, groups also have access to the derived data obtained from the processor, which includes $\mat{W}^{Global}$, $b_{H}^{Global}$, $\mat{M}^{g}$, and $\mu^{Global}$. The central aggregator has access to the latent feature matrices $\mat{H}^{g}$ and bias vectors $b_{H}^{g}$ for items, and the average ratings $\mu^{g}$ from each group. This means that the server does not have access to the raw data of the groups or users. While communicating the aforementioned derivative data, encryption can be used to reduce the risk of an adversary accessing it in transit. When receiving the data, the aggregator assigns a positional identification number to the received data to return each $\mat{M}^{g}$ to the correct group. Because each group trains distinct local CF models locally, the server does not need to store any data after processing and sharing of the results is complete. This removal of data, while minimizing further privacy risks, would require each group to re-participate in training if any of the groups requests a new model. A group may request a new model to utilize newly collected data, or in the case of a user requesting their data to be removed from the system. In the latter case, their raw data (ratings) can just be deleted locally within the group. However, to be GDPR compliant, this user's data should also be removed from the models. This is an open research question in ML systems~\cite{graves2021amnesiac}. Naively, it would require each group to re-participate in federated learning to build new local models.

\subsubsection{Implications of group based design}
\label{subsec:group_implications}
Our proposed method is designed to federate data from groups of users or a set of organizations. The group-based scenario with users, for example, could include Apple's Family Sharing and Netflix profile sharing. This group-based design with multiple users has implications for privacy and communication of the data among the group members. For instance, if the group members use a single device for these systems (household scenario), that device can be considered as the client containing the data of the group and participating in federated learning. However, when the members of the group use multiple devices we need to consider the potential ways of data sharing between the members. The first approach could be that each user's device becomes a client containing data from the other users that are in the same group. This approach would require the members of a group to first aggregate their data among themselves, which has privacy implications for the group. Another option is to take a hierarchical federated learning approach \cite{li2020federated}, where the members of the group send their data to an intermediate server which becomes the client participating in federated learning. While the second option provides better privacy among the group members, the control of the user data is reduced with the addition of an intermediate server. Another implication of the 
groups-based approach is that the method needs the presence of a group, which prevents a single user from participating in federated learning. For example, in our experiments we let the minimum number of group members to be 3 to let each group have an adequate number of ratings. The groups approach also extends to the scenarios where a set of organizations would like to collaborate without sharing their users' data; for example, patients' medical data from multiple collaborating hospitals and student data from collaborating universities. In the case of collaborating organizations, each participant becomes a client similar to the household scenario. Finally, the benefit of group-based design is that the user information is further abstracted, where the server does not need to know which user belongs to which group or how many users are part of the group.

\section{Experiments}
\label{sec:experiments}

In this section we showcase our experiment results and perform an empirical privacy audit of FedSPLIT. We first begin with descriptions of the datasets used in our analysis, and the experimental setup. 

\subsection{Dataset and Experiment Setup}
\label{sec:dataset_and_setup}

\begin{table*}[htb]
\vspace{-0.3em}
\caption{MovieLens 100K and MovieLens 1M datasets statistics.}
 \label{table:dataset_stats}
\begin{tabular}{l|c|c|c|c|c}
\thickhline
\textbf{Dataset}   & \textbf{Users}   & \textbf{Items} & \textbf{Num. Ratings} & \textbf{Sparsity} & \textbf{Mean Num. Groups} \\ \thickhline

MovieLens 100K    & 610   & 9,724 & 100,836    & 0.01 & 36.7 ($\pm$ 2.613) \\
MovieLens 1M      & 6,040 & 3,706 & 1,000,209  & 0.05 & 355.50 ($\pm$ 11.800) \\

\thickhline
\end{tabular}
\vspace{-0.3em}
\end{table*}

We evaluate our method on the MovieLens 100K and MovieLens 1M datasets, popular benchmarking datasets for evaluating recommendation methods \cite{harper2015movielens}. Specifically, we use the explicit feedback of users for movie ratings, which are between 1 and 5, in our analysis. The statistics of these datasets are provided in Table \ref{table:dataset_stats}. For pre-processing, we first remove the users that rated less than 20 movies, then remove the movies that have less than 20 ratings, and finally round up 0.5 ratings to 1. Also, when doing our validation and test splits, we ensure that the users and movies in the training set have at least 1 rating associated with them.

When evaluating FedSPLIT, we randomly assign users to be part of the groups with randomly selected (from a uniform distribution) sizes between 3 and 30. To show that our results are statistically significant, we run our experiments multiple times by splitting the users into groups 10 times for the MovieLens 100k dataset, and 4 times for the bigger MovieLens 1M dataset, where each time we use a distinct random seed. For the non-private CF model baseline (baselines will be explained in Section \ref{sec:baseline_comparisions}), which is not based on groups but rather a single matrix, each distinct experiment with different random seed is used to select a different test set. This way, we report our mean results with a 95\% Confidence Interval (CI). Table \ref{table:dataset_stats} also includes the average number of groups we generate per experiment. 

We test the performance of our method on a held-out test set of size 20\% of that of the entire dataset. For the non-private CF models and each groups' local CF models, the hyper-parameter tuning is performed using validation set sized at 20\% of the training-set with 5 fold cross-validation (CV) per tuning trial using a popular package named \textit{Optuna} \cite{akiba2019optuna}. We run 100 tuning trials for the MovieLens 100K dataset, and 50 tuning trials for the bigger MovieLens 1M dataset. The latent factor matrices for the non-private CF and each groups' local CF are initialized randomly, and the following hyper-parameters are tuned by running the models a maximum of 100 iterations (ranges are given in parentheses): $\alpha$ (0.04-0.08, log-scale), $\beta$ (0.04-0.08, log-scale), $\gamma$ (0.01-0.04, log-scale), $\delta$ (0.01-0.04, log-scale), $\eta_W$ (0.002-0.009, log-scale), $\eta_H$ (0.002-0.009, log-scale), and $k$ (2-$[min(n^g, 21)-1]$). To tune the NMF performed by the server (Section \ref{sec:step_2}), we test for $k \in [2, 4, 6, \cdots, 30]$ with 5 fold CV for each $k$ (again 20\% validation-set size), with maximum 1,000 iterations, and latent factor matrices are initialized with Non-negative Double Singular Value Decomposition (NNDSVD) \cite{boutsidis2008svd}. After the tuning, we run the ultimate CF models for a maximum of 500 iterations using the found optimal hyper-parameters.

\subsection{Performance Analysis}
\label{sec:performance_analysis}

The goal of federated learning is to utilize data from all users to build better performing ML models, while also preserving the privacy of the user data. Therefore, we begin the analysis of FedSPLIT by measuring its improvement of recommendation performance for each groups' model. To do this, we compare the RMSE of distinct local models obtained with FedSPLIT to traditional CF with CNMF for each group across all experiments for both of datasets.

\begin{table*}[htb]
\caption{Numerical summary of Figure \ref{fig:group_SPLIT_impact}.}
 \label{table:result_stats}
 \adjustbox{max width=\textwidth}{%
\begin{tabular}{l|c|c|c|c|c|c}
\thickhline
\textbf{Dataset}   & \textbf{Num. Groups}   & \textbf{Num. Improved} & \textbf{RMSE Improve} & \textbf{RMSE Reduce} & \textbf{Members-RMSE Pearson} & \textbf{Ratings-RMSE Pearson} \\ \thickhline

MovieLens 100K    & 365   & 359 & -0.22 ($\pm$ 0.008)  & 0.07 ($\pm$ 0.092)  & -0.06 & -0.11   \\
MovieLens 1M      & 1,422 & 1,420 & -0.31 ($\pm$ 0.005)  & 0.07  ($\pm$ 0.381) & -0.12 & -0.16 \\

\thickhline
\end{tabular}
}
\vspace{-0.3cm}
\end{table*}

\begin{figure}[t!]
\vspace{-0.01cm}
  \includegraphics[width=\columnwidth]{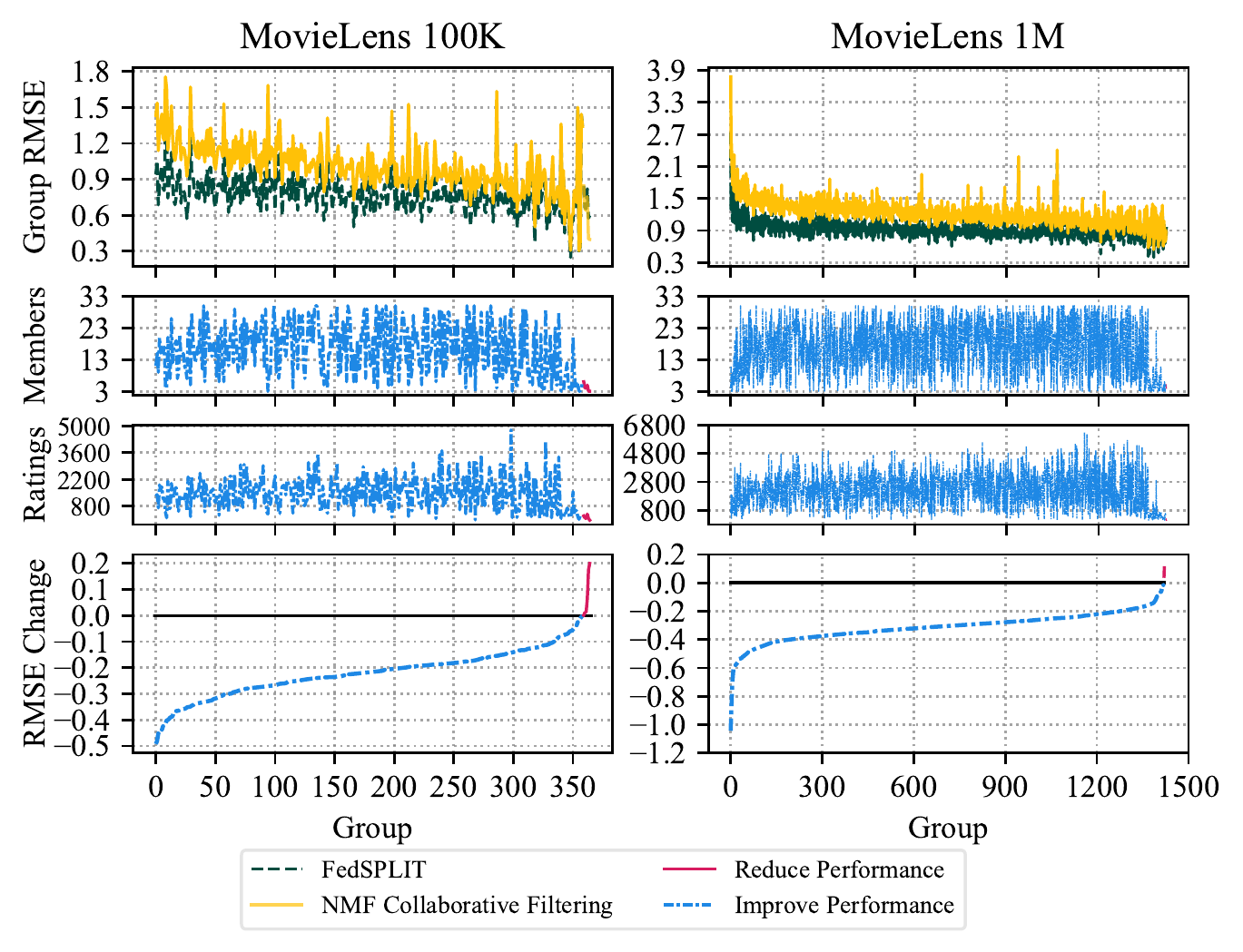}
  \caption{Compare RMSE scores of groups for FedSPLIT and standard CF, and show how much FedSPLIT improves groups' recommendation capability. We also show the number of ratings and members of each group. Results are shown for the groups from all runs of the experiments (10 times for MovieLens 100k, and 5 times for MovieLens 1M).}
  \label{fig:group_SPLIT_impact}
  \vspace{-0.3cm}
\end{figure}

In Figure \ref{fig:group_SPLIT_impact}, we show the RMSE for each group (sorted by the RMSE difference between the local standard CF and the local model after FedSPLIT) obtained using FedSPLIT and standard CF (CNMF), how much group RMSE from CNMF changes after FedSPLIT, and the number of members and ratings belonging to the group. Numerical summary of the figure is also provided in Table \ref{table:result_stats}. The displayed results are for the groups from all runs of the experiments (10 times for MovieLens 100k, and 4 times for MovieLens 1M). 

The first row of Figure \ref{fig:group_SPLIT_impact} shows that FedSPLIT improves the RMSE scores for the majority of the groups for both of the datasets. This can also be seen in the last row of the figure which shows the difference in RMSE between standard CF and FedSPLIT. While a small number of the groups see a drop in performance (increased RMSE) in the MovieLens 100K dataset (6 groups), where the RMSE is increased by 0.07 on average, the remainder of the 365 groups see an average of 0.22 RMSE drop (Table \ref{table:result_stats}). In MovieLens 1M dataset, all but two groups see an average of 0.31 reduction in RMSE (performance improvement). Also, the number of group members and ratings have almost no correlation to the improvement in performance, which is pointed out by fluctuating numbers in rows 2 and 3 in Figure \ref{fig:group_SPLIT_impact}, and the Pearson correlation score in Table \ref{table:result_stats}. We noticed, however, that the groups that saw an increase in RMSE (performance reduction) in the MovieLens 100K dataset, and the groups that saw a small reduction in RMSE (performance improvement) are the ones with a smaller number of members and ratings. Despite this, many other groups with smaller number of members and ratings do see higher improvement in recommendation capabilities. In summary, our results suggest that FedSPLIT can improve the recommendation capability of local models using federated learning.

\subsection{Baseline Comparisons}
\label{sec:baseline_comparisions}

Next, we compare our approach to sate-of-the-art federated recommendation baselines in this section. To the best of our knowledge, FedSPLIT is the first one-shot federated learning recommendation system. Previous works on one-shot federated learning focused on supervised and semi-supervised classification tasks for image and text \cite{Guha2019OneShotFL, Zhou2020DistilledOF, Li2021PracticalOF, yurochkin2019bayesian, kasturi2020fusion, shin2020xor, guha2018model}. Hence, our baselines are federated recommenders with iterative updates. Our goal in this section is to show that our one-shot federated communication approach can perform similar or better to the baseline models that require multi-round client participation.

\begin{table*}[htb]
\caption{Comparison of RMSE scores from both datasets for our one-shot federated method, baseline federated learning models (multi-round communication, sorted by the rounds of communications), and standard CF. Results of our approach, and best performing methods per section are in bold. Not Available (\textit{NA}) is used if the prior work did not provide the particular information. Symbol $\sim$ is used when the information was not numerically reported (ex. reading from a plot), and we report it to the best of our understanding. For CLFM-VFL \cite{zhang2021vertical}, FedRec \cite{lin2020fedrec}, and the baseline using homomorphic encryption \cite{DU2021107700}, we also provide the scores for their lowest reported number of iterations.}
 \label{table:baseline_comparisons}
 \adjustbox{max width=\textwidth}{%
\begin{tabular}{l|c|c|c|c}
\thickhline
                                                      \multicolumn{1}{c}{}&\multicolumn{1}{c}{}&\multicolumn{1}{c}{}&\multicolumn{2}{c}{\textbf{RMSE Results on Datasets}}     \\ 
                                     	                                     	                \cmidrule(lr){4-5}
\multicolumn{1}{l}{\textbf{Method}}    &\multicolumn{1}{c}{\textbf{Reference}}&\multicolumn{1}{c}{\textbf{\# of Comm. Rounds}}& \multicolumn{1}{c}{\textbf{MovieLens 100K}}&\multicolumn{1}{c}{\textbf{MovieLens 1M}}\\ \thickhline
\rowcolor{anti-flashwhite}
\textbf{Standard CF} &&&&\\
\text{Non-Private/Standard CF (CNMF) }                             & -                  & - & \textbf{0.71 ($\pm$ 0.006)}     & \textbf{0.76 ($\pm$ 0.006)}  \\
\text{Groups' Local CF (CNMF) }                                    & -                  & - & 1.00 ($\pm$ 0.022)              & 1.22 ($\pm$ 0.013)  \\
\hline
\rowcolor{anti-flashwhite}
\textbf{Iterative Federated Baselines} &&&&\\
\text{CLFM-VFL}                                                    & \cite{zhang2021vertical} & 1$-$175 & $\sim$3.80 (\textit{NA}) $-\sim$1.00 (\textit{NA})          & \textit{NA} \\
\text{FedRec (SVD++)}                                              & \cite{lin2020fedrec}     & 10$-$100 & $\sim$0.95 (\textit{NA}) $-$ \textbf{0.92 ($\pm$ 0.005)}     & $\sim$0.90 (\textit{NA}) $-$ \textbf{0.84 ($\pm$ 0.001)} \\
\text{Homomorphic Encryption}                                      & \cite{DU2021107700}      & 10$-$100 & $\sim$3.40 (\textit{NA}) $-$ 1.03 (\textit{NA})         & \textit{NA} \\
\text{FCMF}                                                        & \cite{YANG2021106946}    & 50 & 0.95 ($\pm$ 0.005)               & 0.88 ($\pm$ 0.001) \\
\text{FedRecon}                                                    & \cite{Singhal2021FederatedRP}  & 500 & \textit{NA}                      & 0.90 (\textit{NA}) \\
\text{FedGNN}                                                      & \cite{wu2021fedgnn}      & \textit{NA} ($>1$) & \textbf{0.92 (\textit{NA})}      & \textbf{0.84 (\textit{NA})} \\
\text{Two-order FedMMF}                                            & \cite{yang2021practical} & \textit{NA} ($>1$)& \textbf{0.92 ($\pm$ 0.003)}      & \textit{NA} \\
\text{FedMF}                                                       & \cite{chai2020secure, wu2021fedgnn} & \textit{NA} ($>1$) & 0.94 (\textit{NA})               & 0.87 (\textit{NA}) \\ 
\text{FCF}                                                         & \cite{ammad2019federated, wu2021fedgnn}& \textit{NA} ($>1$) & 0.95 (\textit{NA})               & 0.87 (\textit{NA}) \\ 
\hline
\rowcolor{anti-flashwhite}
\textbf{One-shot Federated CF} &&&&\\
\text{FedSPLIT}                                                    & (ours)    & \textbf{1}               & \textbf{0.78 ($\pm$ 0.016)}      & \textbf{0.91 ($\pm$ 0.016)}\\

\thickhline
\end{tabular}
}
\vspace{-0.3em}
\end{table*}

Our baseline models are FedRec with SVD++ \cite{lin2020fedrec} (federated factorization method via batching and stochastic updates), Two-order FedMMF \cite{yang2021practical} (federated masked matrix factorization via secret sharing), FedGNN \cite{wu2021fedgnn} (federated graph neural network), FedMF \cite{chai2020secure} (federated matrix factorization with homomorphic encryption, scores obtained from \cite{wu2021fedgnn}), FCF \cite{ammad2019federated} (first federated CF that uses iterative gradient sharing, scores obtained from \cite{wu2021fedgnn}), FCMF \cite{YANG2021106946} (federated collective matrix factorization for heterogeneous feedback data with differential privacy and homomorphic encryption), CLFM-VFL \cite{zhang2021vertical} (federated learning based on clustering), another homomorphic encryption based method \cite{DU2021107700}, and FedRecon \cite{Singhal2021FederatedRP} (partially local federated learning approach based on meta learning). We also provide results for non-private (standard) CF using CNMF (single matrix with all users), and the results from each group's local CF model prior to obtaining improved local models with FedSPLIT. The baseline comparison results with RMSE scores for both datasets are provided in Table \ref{table:baseline_comparisons}. In this table, the reported RMSE scores for groups' CF, and FedSPLIT is the average of the all groups obtained from each run of the experiments. Also, the reported RMSE scores for the rest of the baselines are extracted from the publication in the ``Reference'' column. Note that for three of our baselines (CLFM-VFL \cite{zhang2021vertical}, FedRec \cite{lin2020fedrec}, and Homomorphic Encryption \cite{DU2021107700}) we have also included their score at the lowest reported iteration (communication round). Since these federated methods are designed to be iterative (i.e. not one-shot), their reported best score require more rounds of iterations.

The results in Table \ref{table:baseline_comparisons}, similar to the results shown in Section \ref{sec:performance_analysis}, show that the groups' recommendation performance improves with FedSPLIT. For MovieLens 100K, before the local models are updated with FedSPLIT, the average RMSE for the groups is 1.00. With FedSPLIT, the average RMSE drops to 0.78. We notice a similar improvement for the MovieLens 1M dataset as well. We can also see that FedSPLIT significantly outperforms the baseline federated recommender systems for MovieLens 100K. On the MovieLens 1M dataset, FedSPLIT performs slightly worse than the baselines. This can be attributed to the more noisy groups with high RMSE scores (see Figure \ref{fig:group_SPLIT_impact}, left side of the plot on row 1 column 2). Since the reported RMSE for our method is the average of all groups, the groups with high RMSE bring up the average score. For example, 30.45\% of the groups achieve RMSE of 0.84 or better (which corresponds to the best baseline on this dataset) with mean RMSE of 0.76. In fact, we can match the mean RMSE of 0.84 if we set RMSE cut-off as 0.96 which corresponds to 68.28\% of the groups. 
To potentially improve the overall performance of our approach and handle the noisy groups, the low-performing groups can be removed from federated learning based on their local models' performance, which we discuss in Section \ref{sec:future_work} as future work. 

Despite our RMSE being the highest for the MovieLens 1M dataset, our RMSE is still close to the baselines, and we achieve this score with a one-shot method while the baselines require multi-round client participation. We can further emphasize the strength of FedSPLIT if we compare the communication cost against the baselines that reported their results per iteration basis (rounds of communication). FedRec (SVD++) needed 100 rounds of communication to achieve RMSE of 0.84 \cite{lin2020fedrec}. The approximate performance of their model at 10 rounds of communication (the lowest they have provided) for MovieLens 100K is RMSE of $\sim$0.95 and RMSE of $\sim$0.90 for MovieLens 1M \cite{lin2020fedrec}. Similarly, homomorphic encryption-based approach reports RMSE of $\sim$3.40 at 10th communication \cite{DU2021107700} and clustering approach reports $\sim$3.8 RMSE at the 1st communication \cite{zhang2021vertical} for MovieLens 100K. Finally, it can be seen that standard CF with CNMF performs significantly better with lower RMSE scores for giving recommendations. However, the standard CF, which relies on gathering raw user data in a centralized server to generate the model, compromises privacy for this level of utility.

\subsection{Initial Privacy Audit}
\label{sec:privacy_audit}

\begin{figure}[htb]
  \includegraphics[width=\columnwidth]{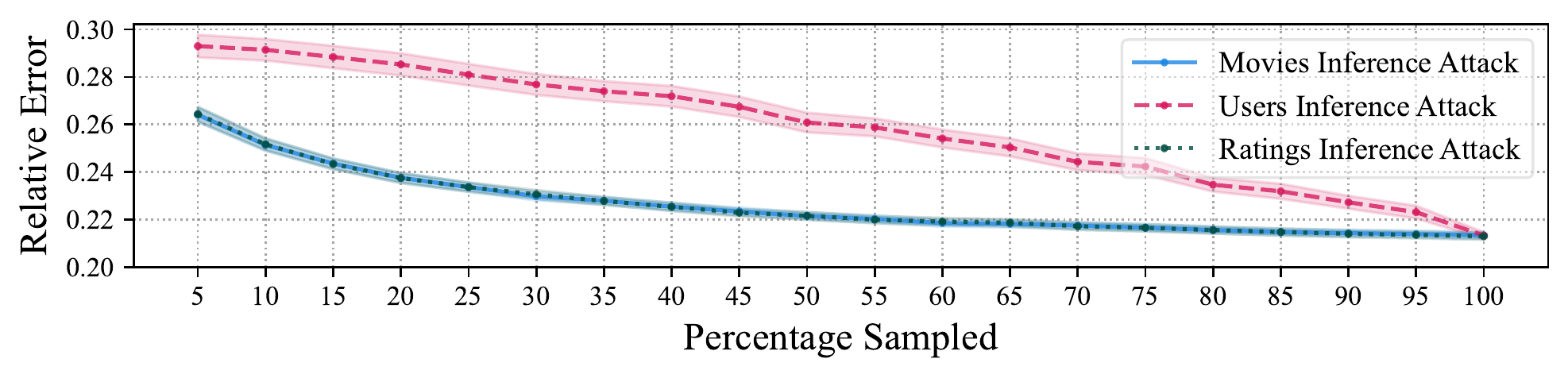}
  \caption{Results for inference attacks based on user, ratings, and movie. Values above 1.0 are plotted separately for scale.} 
  \label{fig:attack_results}
\end{figure}

Motivated by \cite{calandrino2011you}, we analyze the ability of an adversary to learn users' data through inference attacks given levels of auxiliary information on the MovieLens 100k dataset, and present our results in Figure  \ref{fig:attack_results}. FedSPLIT assumes trust at the group level such that the original data $\mat{X}^{g}$, and the derived user patterns $\mat{W}^{g}$ are private. Each group sends $(\mat{H}^{g})^{T}$ and $b_{H}^{g}$ to the central server which also has access to $\mu^{Global}$. This creates an opportunity for an adversary within the group, or anyone who would be able to gain access, to perform a dataset reconstruction attack to derive the raw ratings $\mat{X}^{g}$. Within our threat model, we assume that the adversary has background knowledge of some of the original ratings in $\mat{X}^{g}$ to create $\mat{X}^{g}_{subset}$ containing the subset data. In addition, as mentioned before, FedSPLIT abstracts user information in the context of groups and the shared privacy preserving latent factors are with respect to the items. This means that the adversary may not directly know the number of users present in the group while having access to $(\mat{H}^{g})^{T}$ and $b_{H}^{g}$ alone. Therefore, our adversarial model also assumes that the adversary knows the number of users belonging to the group $g$. Then, the adversary attempts to use $\mat{X}^{g}_{subset}$, $(\mat{H}^{g})^{T}$, $b_{H}^{g}$ and $\mu^{Global}$  to learn a user feature matrix $\mat{W}^{g}_{adv}$ to approximate/reconstruct $\mat{X}^{g}$. This attack is conducted by performing $\mat{W}$-update (Equation~\ref{eqn:masked_W_updates}) and $b_{W}^{g}$-update (Equation~\ref{eqn:bias_W_updates}) optimizing Equation~\ref{eqn:optimize} w.r.t. parameters $\mat{W}^{g}_{{adv}}$ and $b_{W}^{g}$. We measure the success of the adversary's ability to reconstruct the original non-zero review values using relative error (Equation~\ref{eqn:rel_error}) between $\mat{X}^{g}$ and $\mat{X}^{g}_{adv}$ (Equation \ref{eqn:predict_MNMF}). In our experiment, we give the adversary an increasing percentage access (5\% to 100\% with a step size of 5\%) of users within a group (User Inference Attack, rows), of the group's entire movie reviews (Movie Inference Attack, columns), and of movie and user review pairs individually (Ratings Inference Attack, non-zero entries) from $\mat{X}^{g}$. We run our attack against each group obtained from 10 experiment runs (Section \ref{sec:dataset_and_setup}), and report mean relative error with 95\% CI.

As we gain access to entire movie ratings (columns) incrementally, as well as each rating (non-zero entry), we see a near-linear increase in the knowledge gained in reconstruction as shown in Figure \ref{fig:attack_results}. We can attribute the correlated results between the two attacks to columns of $\mat{X}^{g}$ being sparse (i.e., $n^{g}$ users in the group $g$ rated a small-subset of $m$ movies), and thus sampling sparse columns or non-zero entries result in a similar number of non-zeros in $\mat{X}^{g}_{subset}$. Meanwhile, since the number of users $n^{g}$ is significantly smaller than the number of movies $m$ or the number of non-zero entries in $\mat{X}^{g}$, the user-inference attack requires a larger percentage step-size to obtain a unit increase in the sampled number of users which corresponds to increased information (or entries in $\mat{X}^{g}_{subset}$). Overall, for each inference attack, the best result is further away from the perfect reconstruction, even when the adversary has complete knowledge of the data. The inexact reconstruction is attributed to the loss of information in low-rank approximation, which makes the exact reconstruction of the raw data challenging while further enhancing the level of privacy in FedSPLIT.

\section{Future Work}
\label{sec:future_work}

Inspired by the previous one-shot federated learning work that use local model selection to improve the results \cite{Guha2019OneShotFL}, potential areas of future work include carefully selecting the groups that participate in federated learning based on their local performance of the standard CF model or based on the number of members or ratings a group has. In addition, our approach is based on groups, which leaves out single user participants as pointed out in Section \ref{subsec:group_implications}. One possible approach to provide local models to the single users is to perform regression with respect to a randomly initialized $\mat{W}^{g}$ while minimizing $||\mat{X}^{g} - \mat{W}^{g}(\mat{W}^{Global})^T||^{2}_{F}$. This could enable a single user to utilize Equation \ref{eqn:predict_fed_split} without $b_{W}^{g}$ for performing local recommendations. We also leave this for future work. Other future work may include incorporating differential privacy \cite{dwork2008differential} to FedSPLIT to further increase the preservation of individual's privacy.

\section{Conclusion}
\label{sec:conclusion}

In this paper, we have introduced the first approach to recommendation based on one-shot federated learning and collaborative filtering, named FedSPLIT. With FedSPLIT, we obtain client specific recommendations with only a single pair of communications between the server and its clients, after the small initial communication. Our approach utilizes joint non-negative matrix factorization and knowledge distillation. We evaluated our method on two popular recommendation benchmark datasets. Our analysis and results suggest that FedSPLIT improves the predictive capabilities of local standard CF models, and obtains similar or better scores with a one-shot approach compared to the state-of-the-art federated recommendation methods that require multi-round client participation.

\begin{acks}
  This research was partially funded by the Los Alamos National Laboratory (LANL) Laboratory Directed Research and Development (LDRD) grant 20190020DR and LANL Institutional Computing Program, supported by the U.S. Department of Energy National Nuclear Security Administration under Contract No. 89233218CNA000001.
\end{acks}

\bibliographystyle{ACM-Reference-Format}
\bibliography{references}


\begin{thebibliography}{68}


\ifx \showCODEN    \undefined \def \showCODEN     #1{\unskip}     \fi
\ifx \showDOI      \undefined \def \showDOI       #1{#1}\fi
\ifx \showISBNx    \undefined \def \showISBNx     #1{\unskip}     \fi
\ifx \showISBNxiii \undefined \def \showISBNxiii  #1{\unskip}     \fi
\ifx \showISSN     \undefined \def \showISSN      #1{\unskip}     \fi
\ifx \showLCCN     \undefined \def \showLCCN      #1{\unskip}     \fi
\ifx \shownote     \undefined \def \shownote      #1{#1}          \fi
\ifx \showarticletitle \undefined \def \showarticletitle #1{#1}   \fi
\ifx \showURL      \undefined \def \showURL       {\relax}        \fi
\providecommand\bibfield[2]{#2}
\providecommand\bibinfo[2]{#2}
\providecommand\natexlab[1]{#1}
\providecommand\showeprint[2][]{arXiv:#2}

\bibitem[Abadi et~al\mbox{.}(2016)]%
        {abadi2016deep}
\bibfield{author}{\bibinfo{person}{Martin Abadi}, \bibinfo{person}{Andy Chu},
  \bibinfo{person}{Ian Goodfellow}, \bibinfo{person}{H~Brendan McMahan},
  \bibinfo{person}{Ilya Mironov}, \bibinfo{person}{Kunal Talwar}, {and}
  \bibinfo{person}{Li Zhang}.} \bibinfo{year}{2016}\natexlab{}.
\newblock \showarticletitle{Deep learning with differential privacy}. In
  \bibinfo{booktitle}{\emph{ACM SIGSAC conference on computer and
  communications security}}. \bibinfo{pages}{308--318}.
\newblock


\bibitem[Akiba et~al\mbox{.}(2019)]%
        {akiba2019optuna}
\bibfield{author}{\bibinfo{person}{Takuya Akiba}, \bibinfo{person}{Shotaro
  Sano}, \bibinfo{person}{Toshihiko Yanase}, \bibinfo{person}{Takeru Ohta},
  {and} \bibinfo{person}{Masanori Koyama}.} \bibinfo{year}{2019}\natexlab{}.
\newblock \showarticletitle{Optuna: A next-generation hyperparameter
  optimization framework}. In \bibinfo{booktitle}{\emph{25th ACM SIGKDD
  international conference on knowledge discovery \& data mining}}.
  \bibinfo{pages}{2623--2631}.
\newblock


\bibitem[Alsulaimawi(2020)]%
        {alsulaimawi2020non}
\bibfield{author}{\bibinfo{person}{Zahir Alsulaimawi}.}
  \bibinfo{year}{2020}\natexlab{}.
\newblock \showarticletitle{A non-negative matrix factorization framework for
  privacy-preserving and federated learning}. In \bibinfo{booktitle}{\emph{IEEE
  22nd International Workshop on Multimedia Signal Processing (MMSP)}}.
  \bibinfo{pages}{1--6}.
\newblock


\bibitem[Ammad-Ud-Din et~al\mbox{.}(2019)]%
        {ammad2019federated}
\bibfield{author}{\bibinfo{person}{Muhammad Ammad-Ud-Din},
  \bibinfo{person}{Elena Ivannikova}, \bibinfo{person}{Suleiman~A Khan},
  \bibinfo{person}{Were Oyomno}, \bibinfo{person}{Qiang Fu},
  \bibinfo{person}{Kuan~Eeik Tan}, {and} \bibinfo{person}{Adrian Flanagan}.}
  \bibinfo{year}{2019}\natexlab{}.
\newblock \showarticletitle{Federated collaborative filtering for
  privacy-preserving personalized recommendation system}.
\newblock \bibinfo{journal}{\emph{arXiv preprint arXiv:1901.09888}}
  (\bibinfo{year}{2019}).
\newblock


\bibitem[Baltrunas et~al\mbox{.}(2010)]%
        {baltrunas2010group}
\bibfield{author}{\bibinfo{person}{Linas Baltrunas}, \bibinfo{person}{Tadas
  Makcinskas}, {and} \bibinfo{person}{Francesco Ricci}.}
  \bibinfo{year}{2010}\natexlab{}.
\newblock \showarticletitle{Group recommendations with rank aggregation and
  collaborative filtering}. In \bibinfo{booktitle}{\emph{4th ACM conference on
  Recommender systems}}. \bibinfo{pages}{119--126}.
\newblock


\bibitem[Balu and Furon(2016)]%
        {balu2016differentially}
\bibfield{author}{\bibinfo{person}{Raghavendran Balu} {and}
  \bibinfo{person}{Teddy Furon}.} \bibinfo{year}{2016}\natexlab{}.
\newblock \showarticletitle{Differentially private matrix factorization using
  sketching techniques}. In \bibinfo{booktitle}{\emph{4th ACM Workshop on
  Information Hiding and Multimedia Security}}. \bibinfo{pages}{57--62}.
\newblock


\bibitem[Basu(2020)]%
        {basu2020privacy}
\bibfield{author}{\bibinfo{person}{Rahul Basu}.}
  \bibinfo{year}{2020}\natexlab{}.
\newblock \showarticletitle{Privacy-preserving recommendation system using
  federated learning}.
\newblock  (\bibinfo{year}{2020}).
\newblock


\bibitem[Boutsidis and Gallopoulos(2008)]%
        {boutsidis2008svd}
\bibfield{author}{\bibinfo{person}{Christos Boutsidis} {and}
  \bibinfo{person}{Efstratios Gallopoulos}.} \bibinfo{year}{2008}\natexlab{}.
\newblock \showarticletitle{SVD based initialization: A head start for
  nonnegative matrix factorization}.
\newblock \bibinfo{journal}{\emph{Pattern recognition}} \bibinfo{volume}{41},
  \bibinfo{number}{4} (\bibinfo{year}{2008}), \bibinfo{pages}{1350--1362}.
\newblock


\bibitem[Calandrino et~al\mbox{.}(2011)]%
        {calandrino2011you}
\bibfield{author}{\bibinfo{person}{Joseph~A Calandrino}, \bibinfo{person}{Ann
  Kilzer}, \bibinfo{person}{Arvind Narayanan}, \bibinfo{person}{Edward~W
  Felten}, {and} \bibinfo{person}{Vitaly Shmatikov}.}
  \bibinfo{year}{2011}\natexlab{}.
\newblock \showarticletitle{" You might also like:" Privacy risks of
  collaborative filtering}. In \bibinfo{booktitle}{\emph{IEEE symposium on
  security and privacy}}. \bibinfo{pages}{231--246}.
\newblock


\bibitem[Chai et~al\mbox{.}(2020)]%
        {chai2020secure}
\bibfield{author}{\bibinfo{person}{Di Chai}, \bibinfo{person}{Leye Wang},
  \bibinfo{person}{Kai Chen}, {and} \bibinfo{person}{Qiang Yang}.}
  \bibinfo{year}{2020}\natexlab{}.
\newblock \showarticletitle{Secure federated matrix factorization}.
\newblock \bibinfo{journal}{\emph{IEEE Intelligent Systems}}
  \bibinfo{volume}{36}, \bibinfo{number}{5} (\bibinfo{year}{2020}),
  \bibinfo{pages}{11--20}.
\newblock


\bibitem[Chen et~al\mbox{.}(2018)]%
        {chen2018federated}
\bibfield{author}{\bibinfo{person}{Fei Chen}, \bibinfo{person}{Mi Luo},
  \bibinfo{person}{Zhenhua Dong}, \bibinfo{person}{Zhenguo Li}, {and}
  \bibinfo{person}{Xiuqiang He}.} \bibinfo{year}{2018}\natexlab{}.
\newblock \showarticletitle{Federated meta-learning with fast convergence and
  efficient communication}.
\newblock \bibinfo{journal}{\emph{arXiv preprint arXiv:1802.07876}}
  (\bibinfo{year}{2018}).
\newblock


\bibitem[Chen et~al\mbox{.}(2004)]%
        {chen2004impact}
\bibfield{author}{\bibinfo{person}{Pei-Yu Chen}, \bibinfo{person}{Shin-yi Wu},
  {and} \bibinfo{person}{Jungsun Yoon}.} \bibinfo{year}{2004}\natexlab{}.
\newblock \showarticletitle{The impact of online recommendations and consumer
  feedback on sales}.
\newblock  (\bibinfo{year}{2004}).
\newblock


\bibitem[Du et~al\mbox{.}(2021)]%
        {DU2021107700}
\bibfield{author}{\bibinfo{person}{Yongjie Du}, \bibinfo{person}{Deyun Zhou},
  \bibinfo{person}{Yu Xie}, \bibinfo{person}{Jiao Shi}, {and}
  \bibinfo{person}{Maoguo Gong}.} \bibinfo{year}{2021}\natexlab{}.
\newblock \showarticletitle{Federated matrix factorization for
  privacy-preserving recommender systems}.
\newblock \bibinfo{journal}{\emph{Applied Soft Computing}}
  \bibinfo{volume}{111} (\bibinfo{year}{2021}), \bibinfo{pages}{107700}.
\newblock
\showISSN{1568-4946}


\bibitem[Dwork(2008)]%
        {dwork2008differential}
\bibfield{author}{\bibinfo{person}{Cynthia Dwork}.}
  \bibinfo{year}{2008}\natexlab{}.
\newblock \showarticletitle{Differential privacy: A survey of results}. In
  \bibinfo{booktitle}{\emph{International conference on theory and applications
  of models of computation}}. \bibinfo{pages}{1--19}.
\newblock


\bibitem[F{\'e}votte and Cemgil(2009)]%
        {fevotte2009nonnegative}
\bibfield{author}{\bibinfo{person}{C{\'e}dric F{\'e}votte} {and}
  \bibinfo{person}{A~Taylan Cemgil}.} \bibinfo{year}{2009}\natexlab{}.
\newblock \showarticletitle{Nonnegative matrix factorizations as probabilistic
  inference in composite models}. In \bibinfo{booktitle}{\emph{17th European
  Signal Processing Conference}}. \bibinfo{pages}{1913--1917}.
\newblock


\bibitem[Flanagan et~al\mbox{.}(2020)]%
        {flanagan2020federated}
\bibfield{author}{\bibinfo{person}{Adrian Flanagan}, \bibinfo{person}{Were
  Oyomno}, \bibinfo{person}{Alexander Grigorievskiy}, \bibinfo{person}{Kuan~E
  Tan}, \bibinfo{person}{Suleiman~A Khan}, {and} \bibinfo{person}{Muhammad
  Ammad-Ud-Din}.} \bibinfo{year}{2020}\natexlab{}.
\newblock \showarticletitle{Federated multi-view matrix factorization for
  personalized recommendations}. In \bibinfo{booktitle}{\emph{Joint European
  Conference on Machine Learning and Knowledge Discovery in Databases}}.
  \bibinfo{pages}{324--347}.
\newblock


\bibitem[Fredrikson et~al\mbox{.}(2015)]%
        {fredrikson2015model}
\bibfield{author}{\bibinfo{person}{Matt Fredrikson}, \bibinfo{person}{Somesh
  Jha}, {and} \bibinfo{person}{Thomas Ristenpart}.}
  \bibinfo{year}{2015}\natexlab{}.
\newblock \showarticletitle{Model inversion attacks that exploit confidence
  information and basic countermeasures}. In \bibinfo{booktitle}{\emph{22nd ACM
  SIGSAC conference on computer and communications security}}.
  \bibinfo{pages}{1322--1333}.
\newblock


\bibitem[Gao et~al\mbox{.}(2020)]%
        {gao2020privacy}
\bibfield{author}{\bibinfo{person}{Dashan Gao}, \bibinfo{person}{Ben Tan},
  \bibinfo{person}{Ce Ju}, \bibinfo{person}{Vincent~W Zheng}, {and}
  \bibinfo{person}{Qiang Yang}.} \bibinfo{year}{2020}\natexlab{}.
\newblock \showarticletitle{Privacy threats against federated matrix
  factorization}.
\newblock \bibinfo{journal}{\emph{arXiv preprint arXiv:2007.01587}}
  (\bibinfo{year}{2020}).
\newblock


\bibitem[Ghosh et~al\mbox{.}(2020)]%
        {ghosh2020efficient}
\bibfield{author}{\bibinfo{person}{Avishek Ghosh}, \bibinfo{person}{Jichan
  Chung}, \bibinfo{person}{Dong Yin}, {and} \bibinfo{person}{Kannan
  Ramchandran}.} \bibinfo{year}{2020}\natexlab{}.
\newblock \showarticletitle{An efficient framework for clustered federated
  learning}.
\newblock \bibinfo{journal}{\emph{Advances in Neural Information Processing
  Systems}}  \bibinfo{volume}{33} (\bibinfo{year}{2020}),
  \bibinfo{pages}{19586--19597}.
\newblock


\bibitem[Goldberg et~al\mbox{.}(1992)]%
        {goldberg1992using}
\bibfield{author}{\bibinfo{person}{David Goldberg}, \bibinfo{person}{David
  Nichols}, \bibinfo{person}{Brian~M Oki}, {and} \bibinfo{person}{Douglas
  Terry}.} \bibinfo{year}{1992}\natexlab{}.
\newblock \showarticletitle{Using collaborative filtering to weave an
  information tapestry}.
\newblock \bibinfo{journal}{\emph{Commun. ACM}} \bibinfo{volume}{35},
  \bibinfo{number}{12} (\bibinfo{year}{1992}), \bibinfo{pages}{61--70}.
\newblock


\bibitem[Golub et~al\mbox{.}(1999)]%
        {golub1999tikhonov}
\bibfield{author}{\bibinfo{person}{Gene~H Golub},
  \bibinfo{person}{Per~Christian Hansen}, {and} \bibinfo{person}{Dianne~P
  O'Leary}.} \bibinfo{year}{1999}\natexlab{}.
\newblock \showarticletitle{Tikhonov regularization and total least squares}.
\newblock \bibinfo{journal}{\emph{SIAM journal on matrix analysis and
  applications}} \bibinfo{volume}{21}, \bibinfo{number}{1}
  (\bibinfo{year}{1999}), \bibinfo{pages}{185--194}.
\newblock


\bibitem[Gong et~al\mbox{.}(2020)]%
        {gong2020survey}
\bibfield{author}{\bibinfo{person}{Maoguo Gong}, \bibinfo{person}{Yu Xie},
  \bibinfo{person}{Ke Pan}, \bibinfo{person}{Kaiyuan Feng}, {and}
  \bibinfo{person}{Alex~Kai Qin}.} \bibinfo{year}{2020}\natexlab{}.
\newblock \showarticletitle{A survey on differentially private machine
  learning}.
\newblock \bibinfo{journal}{\emph{IEEE computational intelligence magazine}}
  \bibinfo{volume}{15}, \bibinfo{number}{2} (\bibinfo{year}{2020}),
  \bibinfo{pages}{49--64}.
\newblock


\bibitem[Graves et~al\mbox{.}(2021)]%
        {graves2021amnesiac}
\bibfield{author}{\bibinfo{person}{Laura Graves}, \bibinfo{person}{Vineel
  Nagisetty}, {and} \bibinfo{person}{Vijay Ganesh}.}
  \bibinfo{year}{2021}\natexlab{}.
\newblock \showarticletitle{Amnesiac Machine Learning}. In
  \bibinfo{booktitle}{\emph{AAAI Conference on Artificial Intelligence}},
  Vol.~\bibinfo{volume}{35}. \bibinfo{pages}{11516--11524}.
\newblock


\bibitem[Guha and Smith(2018)]%
        {guha2018model}
\bibfield{author}{\bibinfo{person}{Neel Guha} {and} \bibinfo{person}{Virginia
  Smith}.} \bibinfo{year}{2018}\natexlab{}.
\newblock \showarticletitle{Model aggregation via good-enough model spaces}.
\newblock \bibinfo{journal}{\emph{arXiv preprint arXiv:1805.07782}}
  (\bibinfo{year}{2018}).
\newblock


\bibitem[Guha et~al\mbox{.}(2019)]%
        {Guha2019OneShotFL}
\bibfield{author}{\bibinfo{person}{Neel Guha}, \bibinfo{person}{Ameet~S.
  Talwalkar}, {and} \bibinfo{person}{Virginia Smith}.}
  \bibinfo{year}{2019}\natexlab{}.
\newblock \showarticletitle{One-Shot Federated Learning}.
\newblock \bibinfo{journal}{\emph{ArXiv}}  \bibinfo{volume}{abs/1902.11175}
  (\bibinfo{year}{2019}).
\newblock


\bibitem[Harper and Konstan(2015)]%
        {harper2015movielens}
\bibfield{author}{\bibinfo{person}{F~Maxwell Harper} {and}
  \bibinfo{person}{Joseph~A Konstan}.} \bibinfo{year}{2015}\natexlab{}.
\newblock \showarticletitle{The movielens datasets: History and context}.
\newblock \bibinfo{journal}{\emph{Acm transactions on interactive intelligent
  systems (tiis)}} \bibinfo{volume}{5}, \bibinfo{number}{4}
  (\bibinfo{year}{2015}), \bibinfo{pages}{1--19}.
\newblock


\bibitem[Hug(2020)]%
        {Hug2020}
\bibfield{author}{\bibinfo{person}{Nicolas Hug}.}
  \bibinfo{year}{2020}\natexlab{}.
\newblock \showarticletitle{Surprise: A Python library for recommender
  systems}.
\newblock \bibinfo{journal}{\emph{Journal of Open Source Software}}
  \bibinfo{volume}{5}, \bibinfo{number}{52} (\bibinfo{year}{2020}),
  \bibinfo{pages}{2174}.
\newblock
\urldef\tempurl%
\url{https://doi.org/10.21105/joss.02174}
\showDOI{\tempurl}


\bibitem[Jalalirad et~al\mbox{.}(2019)]%
        {jalalirad2019simple}
\bibfield{author}{\bibinfo{person}{Amir Jalalirad}, \bibinfo{person}{Marco
  Scavuzzo}, \bibinfo{person}{Catalin Capota}, {and} \bibinfo{person}{Michael
  Sprague}.} \bibinfo{year}{2019}\natexlab{}.
\newblock \showarticletitle{A simple and efficient federated recommender
  system}. In \bibinfo{booktitle}{\emph{6th IEEE/ACM International Conference
  on Big Data Computing, Applications and Technologies}}.
  \bibinfo{pages}{53--58}.
\newblock


\bibitem[Kasturi et~al\mbox{.}(2020)]%
        {kasturi2020fusion}
\bibfield{author}{\bibinfo{person}{Anirudh Kasturi},
  \bibinfo{person}{Anish~Reddy Ellore}, {and} \bibinfo{person}{Chittaranjan
  Hota}.} \bibinfo{year}{2020}\natexlab{}.
\newblock \showarticletitle{Fusion learning: A one shot federated learning}. In
  \bibinfo{booktitle}{\emph{International Conference on Computational
  Science}}. \bibinfo{pages}{424--436}.
\newblock


\bibitem[Khan et~al\mbox{.}(2021)]%
        {khan2021payload}
\bibfield{author}{\bibinfo{person}{Farwa~K Khan}, \bibinfo{person}{Adrian
  Flanagan}, \bibinfo{person}{Kuan~Eeik Tan}, \bibinfo{person}{Zareen Alamgir},
  {and} \bibinfo{person}{Muhammad Ammad-Ud-Din}.}
  \bibinfo{year}{2021}\natexlab{}.
\newblock \showarticletitle{A Payload Optimization Method for Federated
  Recommender Systems}. In \bibinfo{booktitle}{\emph{15th ACM Conference on
  Recommender Systems}}. \bibinfo{pages}{432--442}.
\newblock


\bibitem[Kim et~al\mbox{.}(2016)]%
        {kim2016efficient}
\bibfield{author}{\bibinfo{person}{Sungwook Kim}, \bibinfo{person}{Jinsu Kim},
  \bibinfo{person}{Dongyoung Koo}, \bibinfo{person}{Yuna Kim},
  \bibinfo{person}{Hyunsoo Yoon}, {and} \bibinfo{person}{Junbum Shin}.}
  \bibinfo{year}{2016}\natexlab{}.
\newblock \showarticletitle{Efficient privacy-preserving matrix factorization
  via fully homomorphic encryption}. In \bibinfo{booktitle}{\emph{11th ACM on
  Asia Conference on Computer and Communications Security}}.
  \bibinfo{pages}{617--628}.
\newblock


\bibitem[Kone{\v{c}}n{\`y} et~al\mbox{.}(2015)]%
        {konevcny2015federated}
\bibfield{author}{\bibinfo{person}{Jakub Kone{\v{c}}n{\`y}},
  \bibinfo{person}{Brendan McMahan}, {and} \bibinfo{person}{Daniel Ramage}.}
  \bibinfo{year}{2015}\natexlab{}.
\newblock \showarticletitle{Federated optimization: Distributed optimization
  beyond the datacenter}.
\newblock \bibinfo{journal}{\emph{arXiv preprint arXiv:1511.03575}}
  (\bibinfo{year}{2015}).
\newblock


\bibitem[Kone{\v{c}}n{\`y} et~al\mbox{.}(2016)]%
        {konevcny2016federated}
\bibfield{author}{\bibinfo{person}{Jakub Kone{\v{c}}n{\`y}},
  \bibinfo{person}{H~Brendan McMahan}, \bibinfo{person}{Felix~X Yu},
  \bibinfo{person}{Peter Richt{\'a}rik}, \bibinfo{person}{Ananda~Theertha
  Suresh}, {and} \bibinfo{person}{Dave Bacon}.}
  \bibinfo{year}{2016}\natexlab{}.
\newblock \showarticletitle{Federated learning: Strategies for improving
  communication efficiency}.
\newblock \bibinfo{journal}{\emph{arXiv preprint arXiv:1610.05492}}
  (\bibinfo{year}{2016}).
\newblock


\bibitem[Koren et~al\mbox{.}(2009)]%
        {koren2009matrix}
\bibfield{author}{\bibinfo{person}{Yehuda Koren}, \bibinfo{person}{Robert
  Bell}, {and} \bibinfo{person}{Chris Volinsky}.}
  \bibinfo{year}{2009}\natexlab{}.
\newblock \showarticletitle{Matrix factorization techniques for recommender
  systems}.
\newblock \bibinfo{journal}{\emph{Computer}} \bibinfo{volume}{42},
  \bibinfo{number}{8} (\bibinfo{year}{2009}), \bibinfo{pages}{30--37}.
\newblock


\bibitem[Krebs et~al\mbox{.}(2019)]%
        {krebs2019tell}
\bibfield{author}{\bibinfo{person}{Luciana~Monteiro Krebs},
  \bibinfo{person}{Oscar~Luis Alvarado~Rodriguez}, \bibinfo{person}{Pierre
  Dewitte}, \bibinfo{person}{Jef Ausloos}, \bibinfo{person}{David Geerts},
  \bibinfo{person}{Laurens Naudts}, {and} \bibinfo{person}{Katrien Verbert}.}
  \bibinfo{year}{2019}\natexlab{}.
\newblock \showarticletitle{Tell me what you know: GDPR implications on
  designing transparency and accountability for news recommender systems}. In
  \bibinfo{booktitle}{\emph{Extended Abstracts of the 2019 CHI Conference on
  Human Factors in Computing Systems}}. \bibinfo{pages}{1--6}.
\newblock


\bibitem[Lee and Seung(1999)]%
        {lee1999learning}
\bibfield{author}{\bibinfo{person}{Daniel~D Lee} {and}
  \bibinfo{person}{H~Sebastian Seung}.} \bibinfo{year}{1999}\natexlab{}.
\newblock \showarticletitle{Learning the parts of objects by non-negative
  matrix factorization}.
\newblock \bibinfo{journal}{\emph{Nature}} \bibinfo{volume}{401},
  \bibinfo{number}{6755} (\bibinfo{year}{1999}), \bibinfo{pages}{788--791}.
\newblock


\bibitem[Li et~al\mbox{.}(2021)]%
        {Li2021PracticalOF}
\bibfield{author}{\bibinfo{person}{Qinbin Li}, \bibinfo{person}{Bingsheng He},
  {and} \bibinfo{person}{Dawn~Xiaodong Song}.} \bibinfo{year}{2021}\natexlab{}.
\newblock \showarticletitle{Practical One-Shot Federated Learning for
  Cross-Silo Setting}. In \bibinfo{booktitle}{\emph{IJCAI}}.
\newblock


\bibitem[Li et~al\mbox{.}(2020)]%
        {li2020federated}
\bibfield{author}{\bibinfo{person}{Tian Li}, \bibinfo{person}{Anit~Kumar Sahu},
  \bibinfo{person}{Ameet Talwalkar}, {and} \bibinfo{person}{Virginia Smith}.}
  \bibinfo{year}{2020}\natexlab{}.
\newblock \showarticletitle{Federated learning: Challenges, methods, and future
  directions}.
\newblock \bibinfo{journal}{\emph{IEEE Signal Processing Magazine}}
  \bibinfo{volume}{37}, \bibinfo{number}{3} (\bibinfo{year}{2020}),
  \bibinfo{pages}{50--60}.
\newblock


\bibitem[Lin et~al\mbox{.}(2020)]%
        {lin2020fedrec}
\bibfield{author}{\bibinfo{person}{Guanyu Lin}, \bibinfo{person}{Feng Liang},
  \bibinfo{person}{Weike Pan}, {and} \bibinfo{person}{Zhong Ming}.}
  \bibinfo{year}{2020}\natexlab{}.
\newblock \showarticletitle{Fedrec: Federated recommendation with explicit
  feedback}.
\newblock \bibinfo{journal}{\emph{IEEE Intelligent Systems}}
  \bibinfo{volume}{36}, \bibinfo{number}{5} (\bibinfo{year}{2020}),
  \bibinfo{pages}{21--30}.
\newblock


\bibitem[Liu et~al\mbox{.}(2022)]%
        {liu2022threats}
\bibfield{author}{\bibinfo{person}{Pengrui Liu}, \bibinfo{person}{Xiangrui Xu},
  {and} \bibinfo{person}{Wei Wang}.} \bibinfo{year}{2022}\natexlab{}.
\newblock \showarticletitle{Threats, attacks and defenses to federated
  learning: issues, taxonomy and perspectives}.
\newblock \bibinfo{journal}{\emph{Cybersecurity}} \bibinfo{volume}{5},
  \bibinfo{number}{1} (\bibinfo{year}{2022}), \bibinfo{pages}{1--19}.
\newblock


\bibitem[L{\"u} et~al\mbox{.}(2012)]%
        {lu2012recommender}
\bibfield{author}{\bibinfo{person}{Linyuan L{\"u}},
  \bibinfo{person}{Mat{\'u}{\v{s}} Medo}, \bibinfo{person}{Chi~Ho Yeung},
  \bibinfo{person}{Yi-Cheng Zhang}, \bibinfo{person}{Zi-Ke Zhang}, {and}
  \bibinfo{person}{Tao Zhou}.} \bibinfo{year}{2012}\natexlab{}.
\newblock \showarticletitle{Recommender systems}.
\newblock \bibinfo{journal}{\emph{Physics reports}} \bibinfo{volume}{519},
  \bibinfo{number}{1} (\bibinfo{year}{2012}), \bibinfo{pages}{1--49}.
\newblock


\bibitem[Luping et~al\mbox{.}(2019)]%
        {luping2019cmfl}
\bibfield{author}{\bibinfo{person}{Wang Luping}, \bibinfo{person}{Wang Wei},
  {and} \bibinfo{person}{LI Bo}.} \bibinfo{year}{2019}\natexlab{}.
\newblock \showarticletitle{CMFL: Mitigating communication overhead for
  federated learning}. In \bibinfo{booktitle}{\emph{IEEE 39th international
  conference on distributed computing systems (ICDCS)}}.
  \bibinfo{pages}{954--964}.
\newblock


\bibitem[McMahan et~al\mbox{.}(2017)]%
        {mcmahan2017communication}
\bibfield{author}{\bibinfo{person}{Brendan McMahan}, \bibinfo{person}{Eider
  Moore}, \bibinfo{person}{Daniel Ramage}, \bibinfo{person}{Seth Hampson},
  {and} \bibinfo{person}{Blaise~Aguera y Arcas}.}
  \bibinfo{year}{2017}\natexlab{}.
\newblock \showarticletitle{Communication-efficient learning of deep networks
  from decentralized data}. In \bibinfo{booktitle}{\emph{Artificial
  intelligence and statistics}}. PMLR, \bibinfo{pages}{1273--1282}.
\newblock


\bibitem[McMahan et~al\mbox{.}(2016)]%
        {mcmahan2016federated}
\bibfield{author}{\bibinfo{person}{H~Brendan McMahan}, \bibinfo{person}{Eider
  Moore}, \bibinfo{person}{Daniel Ramage}, {and}
  \bibinfo{person}{Blaise~Ag{\"u}era y Arcas}.}
  \bibinfo{year}{2016}\natexlab{}.
\newblock \showarticletitle{Federated learning of deep networks using model
  averaging}.
\newblock \bibinfo{journal}{\emph{arXiv preprint arXiv:1602.05629}}
  (\bibinfo{year}{2016}).
\newblock


\bibitem[Milano et~al\mbox{.}(2020)]%
        {milano2020recommender}
\bibfield{author}{\bibinfo{person}{Silvia Milano},
  \bibinfo{person}{Mariarosaria Taddeo}, {and} \bibinfo{person}{Luciano
  Floridi}.} \bibinfo{year}{2020}\natexlab{}.
\newblock \showarticletitle{Recommender systems and their ethical challenges}.
\newblock \bibinfo{journal}{\emph{Ai \& Society}} \bibinfo{volume}{35},
  \bibinfo{number}{4} (\bibinfo{year}{2020}), \bibinfo{pages}{957--967}.
\newblock


\bibitem[Minto et~al\mbox{.}(2021)]%
        {10.1145/3460231.3474262}
\bibfield{author}{\bibinfo{person}{Lorenzo Minto}, \bibinfo{person}{Moritz
  Haller}, \bibinfo{person}{Benjamin Livshits}, {and} \bibinfo{person}{Hamed
  Haddadi}.} \bibinfo{year}{2021}\natexlab{}.
\newblock \bibinfo{booktitle}{\emph{Stronger Privacy for Federated
  Collaborative Filtering With Implicit Feedback}}.
\newblock \bibinfo{pages}{342--350}.
\newblock
\showISBNx{9781450384582}


\bibitem[O’connor et~al\mbox{.}(2001)]%
        {o2001polylens}
\bibfield{author}{\bibinfo{person}{Mark O’connor}, \bibinfo{person}{Dan
  Cosley}, \bibinfo{person}{Joseph~A Konstan}, {and} \bibinfo{person}{John
  Riedl}.} \bibinfo{year}{2001}\natexlab{}.
\newblock \showarticletitle{PolyLens: A recommender system for groups of
  users}. In \bibinfo{booktitle}{\emph{ECSCW 2001}}. \bibinfo{pages}{199--218}.
\newblock


\bibitem[Resnick and Varian(1997)]%
        {resnick1997recommender}
\bibfield{author}{\bibinfo{person}{Paul Resnick} {and} \bibinfo{person}{Hal~R
  Varian}.} \bibinfo{year}{1997}\natexlab{}.
\newblock \showarticletitle{Recommender systems}.
\newblock \bibinfo{journal}{\emph{Commun. ACM}} \bibinfo{volume}{40},
  \bibinfo{number}{3} (\bibinfo{year}{1997}), \bibinfo{pages}{56--58}.
\newblock


\bibitem[Shin et~al\mbox{.}(2020)]%
        {shin2020xor}
\bibfield{author}{\bibinfo{person}{MyungJae Shin}, \bibinfo{person}{Chihoon
  Hwang}, \bibinfo{person}{Joongheon Kim}, \bibinfo{person}{Jihong Park},
  \bibinfo{person}{Mehdi Bennis}, {and} \bibinfo{person}{Seong-Lyun Kim}.}
  \bibinfo{year}{2020}\natexlab{}.
\newblock \showarticletitle{Xor mixup: Privacy-preserving data augmentation for
  one-shot federated learning}.
\newblock \bibinfo{journal}{\emph{arXiv preprint arXiv:2006.05148}}
  (\bibinfo{year}{2020}).
\newblock


\bibitem[Shokri et~al\mbox{.}(2017)]%
        {shokri2017membership}
\bibfield{author}{\bibinfo{person}{Reza Shokri}, \bibinfo{person}{Marco
  Stronati}, \bibinfo{person}{Congzheng Song}, {and} \bibinfo{person}{Vitaly
  Shmatikov}.} \bibinfo{year}{2017}\natexlab{}.
\newblock \showarticletitle{Membership inference attacks against machine
  learning models}. In \bibinfo{booktitle}{\emph{IEEE symposium on security and
  privacy (SP)}}. \bibinfo{pages}{3--18}.
\newblock


\bibitem[Singhal et~al\mbox{.}(2021a)]%
        {singhal2021federated}
\bibfield{author}{\bibinfo{person}{Karan Singhal}, \bibinfo{person}{Hakim
  Sidahmed}, \bibinfo{person}{Zachary Garrett}, \bibinfo{person}{Shanshan Wu},
  \bibinfo{person}{John Rush}, {and} \bibinfo{person}{Sushant Prakash}.}
  \bibinfo{year}{2021}\natexlab{a}.
\newblock \showarticletitle{Federated reconstruction: Partially local federated
  learning}.
\newblock \bibinfo{journal}{\emph{Advances in Neural Information Processing
  Systems}}  \bibinfo{volume}{34} (\bibinfo{year}{2021}).
\newblock


\bibitem[Singhal et~al\mbox{.}(2021b)]%
        {Singhal2021FederatedRP}
\bibfield{author}{\bibinfo{person}{K. Singhal}, \bibinfo{person}{Hakim
  Sidahmed}, \bibinfo{person}{Zachary Garrett}, \bibinfo{person}{Shanshan Wu},
  \bibinfo{person}{Keith Rush}, {and} \bibinfo{person}{Sushant Prakash}.}
  \bibinfo{year}{2021}\natexlab{b}.
\newblock \showarticletitle{Federated Reconstruction: Partially Local Federated
  Learning}.
\newblock \bibinfo{journal}{\emph{ArXiv}}  \bibinfo{volume}{abs/2102.03448}
  (\bibinfo{year}{2021}).
\newblock


\bibitem[Smith et~al\mbox{.}(2017)]%
        {smith2017federated}
\bibfield{author}{\bibinfo{person}{Virginia Smith}, \bibinfo{person}{Chao-Kai
  Chiang}, \bibinfo{person}{Maziar Sanjabi}, {and} \bibinfo{person}{Ameet~S
  Talwalkar}.} \bibinfo{year}{2017}\natexlab{}.
\newblock \showarticletitle{Federated multi-task learning}.
\newblock \bibinfo{journal}{\emph{Advances in neural information processing
  systems}}  \bibinfo{volume}{30} (\bibinfo{year}{2017}).
\newblock


\bibitem[Su and Khoshgoftaar(2009)]%
        {su2009survey}
\bibfield{author}{\bibinfo{person}{Xiaoyuan Su} {and} \bibinfo{person}{Taghi~M
  Khoshgoftaar}.} \bibinfo{year}{2009}\natexlab{}.
\newblock \showarticletitle{A survey of collaborative filtering techniques}.
\newblock \bibinfo{journal}{\emph{Advances in artificial intelligence}}
  \bibinfo{volume}{2009} (\bibinfo{year}{2009}).
\newblock


\bibitem[Sun et~al\mbox{.}(2018)]%
        {sun2018private}
\bibfield{author}{\bibinfo{person}{Xiaoqiang Sun}, \bibinfo{person}{Peng
  Zhang}, \bibinfo{person}{Joseph~K Liu}, \bibinfo{person}{Jianping Yu}, {and}
  \bibinfo{person}{Weixin Xie}.} \bibinfo{year}{2018}\natexlab{}.
\newblock \showarticletitle{Private machine learning classification based on
  fully homomorphic encryption}.
\newblock \bibinfo{journal}{\emph{IEEE Transactions on Emerging Topics in
  Computing}} \bibinfo{volume}{8}, \bibinfo{number}{2} (\bibinfo{year}{2018}),
  \bibinfo{pages}{352--364}.
\newblock


\bibitem[Tan et~al\mbox{.}(2020)]%
        {tan2020federated}
\bibfield{author}{\bibinfo{person}{Ben Tan}, \bibinfo{person}{Bo Liu},
  \bibinfo{person}{Vincent Zheng}, {and} \bibinfo{person}{Qiang Yang}.}
  \bibinfo{year}{2020}\natexlab{}.
\newblock \showarticletitle{A federated recommender system for online
  services}. In \bibinfo{booktitle}{\emph{14th ACM Conference on Recommender
  Systems}}. \bibinfo{pages}{579--581}.
\newblock


\bibitem[Vallet et~al\mbox{.}(2014)]%
        {vallet2014matrix}
\bibfield{author}{\bibinfo{person}{David Vallet}, \bibinfo{person}{Arik
  Friedman}, {and} \bibinfo{person}{Shlomo Berkovsky}.}
  \bibinfo{year}{2014}\natexlab{}.
\newblock \showarticletitle{Matrix factorization without user data retention}.
  In \bibinfo{booktitle}{\emph{Pacific-Asia Conference on Knowledge Discovery
  and Data Mining}}. \bibinfo{pages}{569--580}.
\newblock


\bibitem[Wang et~al\mbox{.}(2019)]%
        {wang2019matcha}
\bibfield{author}{\bibinfo{person}{Jianyu Wang}, \bibinfo{person}{Anit~Kumar
  Sahu}, \bibinfo{person}{Zhouyi Yang}, \bibinfo{person}{Gauri Joshi}, {and}
  \bibinfo{person}{Soummya Kar}.} \bibinfo{year}{2019}\natexlab{}.
\newblock \showarticletitle{Matcha: Speeding up decentralized sgd via matching
  decomposition sampling}. In \bibinfo{booktitle}{\emph{6th Indian Control
  Conference (ICC)}}. \bibinfo{pages}{299--300}.
\newblock


\bibitem[Wu et~al\mbox{.}(2021)]%
        {wu2021fedgnn}
\bibfield{author}{\bibinfo{person}{Chuhan Wu}, \bibinfo{person}{Fangzhao Wu},
  \bibinfo{person}{Yang Cao}, \bibinfo{person}{Yongfeng Huang}, {and}
  \bibinfo{person}{Xing Xie}.} \bibinfo{year}{2021}\natexlab{}.
\newblock \showarticletitle{Fedgnn: Federated graph neural network for
  privacy-preserving recommendation}.
\newblock \bibinfo{journal}{\emph{arXiv preprint arXiv:2102.04925}}
  (\bibinfo{year}{2021}).
\newblock


\bibitem[Xu et~al\mbox{.}(2012)]%
        {xu2012alternating}
\bibfield{author}{\bibinfo{person}{Yangyang Xu}, \bibinfo{person}{Wotao Yin},
  \bibinfo{person}{Zaiwen Wen}, {and} \bibinfo{person}{Yin Zhang}.}
  \bibinfo{year}{2012}\natexlab{}.
\newblock \showarticletitle{An alternating direction algorithm for matrix
  completion with nonnegative factors}.
\newblock \bibinfo{journal}{\emph{Frontiers of Mathematics in China}}
  \bibinfo{volume}{7}, \bibinfo{number}{2} (\bibinfo{year}{2012}),
  \bibinfo{pages}{365--384}.
\newblock


\bibitem[Yang et~al\mbox{.}(2021a)]%
        {YANG2021106946}
\bibfield{author}{\bibinfo{person}{Enyue Yang}, \bibinfo{person}{Yunfeng
  Huang}, \bibinfo{person}{Feng Liang}, \bibinfo{person}{Weike Pan}, {and}
  \bibinfo{person}{Zhong Ming}.} \bibinfo{year}{2021}\natexlab{a}.
\newblock \showarticletitle{FCMF: Federated collective matrix factorization for
  heterogeneous collaborative filtering}.
\newblock \bibinfo{journal}{\emph{Knowledge-Based Systems}}
  \bibinfo{volume}{220} (\bibinfo{year}{2021}), \bibinfo{pages}{106946}.
\newblock
\showISSN{0950-7051}


\bibitem[Yang et~al\mbox{.}(2021b)]%
        {yang2021practical}
\bibfield{author}{\bibinfo{person}{Liu Yang}, \bibinfo{person}{Ben Tan},
  \bibinfo{person}{Bo Liu}, \bibinfo{person}{Vincent~W Zheng},
  \bibinfo{person}{Kai Chen}, {and} \bibinfo{person}{Qiang Yang}.}
  \bibinfo{year}{2021}\natexlab{b}.
\newblock \showarticletitle{Practical and Secure Federated Recommendation with
  Personalized Masks}.
\newblock \bibinfo{journal}{\emph{arXiv preprint arXiv:2109.02464}}
  (\bibinfo{year}{2021}).
\newblock


\bibitem[Yang et~al\mbox{.}(2020)]%
        {yang2020federated}
\bibfield{author}{\bibinfo{person}{Liu Yang}, \bibinfo{person}{Ben Tan},
  \bibinfo{person}{Vincent~W Zheng}, \bibinfo{person}{Kai Chen}, {and}
  \bibinfo{person}{Qiang Yang}.} \bibinfo{year}{2020}\natexlab{}.
\newblock \showarticletitle{Federated recommendation systems}.
\newblock In \bibinfo{booktitle}{\emph{Federated Learning}}.
  \bibinfo{pages}{225--239}.
\newblock


\bibitem[Ying(2020)]%
        {ying2020shared}
\bibfield{author}{\bibinfo{person}{Senci Ying}.}
  \bibinfo{year}{2020}\natexlab{}.
\newblock \showarticletitle{Shared MF: A privacy-preserving recommendation
  system}.
\newblock \bibinfo{journal}{\emph{arXiv preprint arXiv:2008.07759}}
  (\bibinfo{year}{2020}).
\newblock


\bibitem[Yurochkin et~al\mbox{.}(2019)]%
        {yurochkin2019bayesian}
\bibfield{author}{\bibinfo{person}{Mikhail Yurochkin}, \bibinfo{person}{Mayank
  Agarwal}, \bibinfo{person}{Soumya Ghosh}, \bibinfo{person}{Kristjan
  Greenewald}, \bibinfo{person}{Nghia Hoang}, {and} \bibinfo{person}{Yasaman
  Khazaeni}.} \bibinfo{year}{2019}\natexlab{}.
\newblock \showarticletitle{Bayesian nonparametric federated learning of neural
  networks}. In \bibinfo{booktitle}{\emph{International Conference on Machine
  Learning}}. \bibinfo{pages}{7252--7261}.
\newblock


\bibitem[Zhang and Jiang(2021)]%
        {zhang2021vertical}
\bibfield{author}{\bibinfo{person}{JianFei Zhang} {and} \bibinfo{person}{YuChen
  Jiang}.} \bibinfo{year}{2021}\natexlab{}.
\newblock \showarticletitle{A vertical federation recommendation method based
  on clustering and latent factor model}. In
  \bibinfo{booktitle}{\emph{International Conference on Electronic Information
  Engineering and Computer Science (EIECS)}}. \bibinfo{pages}{362--366}.
\newblock


\bibitem[Zhou et~al\mbox{.}(2020)]%
        {Zhou2020DistilledOF}
\bibfield{author}{\bibinfo{person}{Yanlin Zhou}, \bibinfo{person}{George Pu},
  \bibinfo{person}{Xiyao Ma}, \bibinfo{person}{Xiaolin Li}, {and}
  \bibinfo{person}{Dapeng~Oliver Wu}.} \bibinfo{year}{2020}\natexlab{}.
\newblock \showarticletitle{Distilled One-Shot Federated Learning}.
\newblock \bibinfo{journal}{\emph{ArXiv}}  \bibinfo{volume}{abs/2009.07999}
  (\bibinfo{year}{2020}).
\newblock


\bibitem[Zhu et~al\mbox{.}(2019)]%
        {zhu2019deep}
\bibfield{author}{\bibinfo{person}{Ligeng Zhu}, \bibinfo{person}{Zhijian Liu},
  {and} \bibinfo{person}{Song Han}.} \bibinfo{year}{2019}\natexlab{}.
\newblock \showarticletitle{Deep leakage from gradients}.
\newblock \bibinfo{journal}{\emph{Advances in Neural Information Processing
  Systems}}  \bibinfo{volume}{32} (\bibinfo{year}{2019}).
\newblock


\end{thebibliography}

\end{document}